\documentclass{article}

\usepackage[preprint]{neurips_2024}

\usepackage[utf8]{inputenc} 
\usepackage[T1]{fontenc}    
\usepackage{hyperref}       
\usepackage{url}            
\usepackage{booktabs}       
\usepackage{amsfonts}       
\usepackage{nicefrac}       
\usepackage{microtype}      
\usepackage{xcolor}         
\usepackage[accsupp]{axessibility}
\usepackage{times}
\usepackage{graphicx,xcolor}
\usepackage{float}
\usepackage{subcaption}
\usepackage{amsmath,amssymb,bm,mathtools}
\usepackage{algorithm,algpseudocode}
\usepackage{dcolumn,colortbl,adjustbox,multirow,setspace}
\usepackage{cleveref}
\usepackage{natbib}       
\graphicspath{{figures/}}

\newcommand{\supddagger}{\textsuperscript{\textdaggerdbl}}
\makeatletter%
\newcolumntype{L}{D{.}{.}{2,2}}
\newcolumntype{B}[3]{>{\boldmath\DC@{#1}{#2}{#3}}c<{\DC@end}}
\newcommand{\tbnum}[1]{\multicolumn{1}{B{.}{.}{2,2}}{#1}}

\newcommand{\thead}[1]{\multicolumn{1}{c}{\textbf{#1}}}

\makeatother%

\newcommand{\mnist}{MNIST}
\newcommand{\cifarx}{CIFAR-10}
\newcommand{\imagenet}{ImageNet}
\newcommand{\cifarc}{CIFAR-100}

\newcommand\yyrloss{{{MIFPE}}}
\newcommand\yuloss{{\text{T-MIFPE}}}

\newcommand{\tbgreen}[1]{\textcolor[rgb]{0,0.502,0}{#1}}

\newcommand{\momentum}{\nu}

\DeclareMathOperator{\sceloss}{%
    \mathcal{L}^\mathrm{ce}}

\DeclareMathOperator{\cwloss}{\mathcal{L}^\mathrm{cw}}
\DeclareMathOperator{\dlrloss}{\mathcal{L}^\mathrm{dlr}}
\DeclareMathOperator{\lnsceloss}{\mathcal{L}^\mathrm{MIFPE}}
\DeclareMathOperator{\lnscelossyu}{\mathcal{L}^\text{T-MIFPE}}

\title{Theoretical Analysis of Relative Errors in Gradient Computations for Adversarial Attacks with CE Loss}

\author{%
  Yunrui Yu \\
  Postdoc \\
  Tsinghua University \\
  \texttt{yuyunrui@mail.tsinghua.edu.cn} \\
  \And
  Hang Su \\
  Associate Professor \\
  Computer Science, Tsinghua University \\
  \texttt{hangsu@mail.tsinghua.edu.cn} \\
  \And
  Cheng-zhong Xu \\
  Full Professor \\
  University of Macau \\
  \texttt{chengzhongxu@um.edu.mo} \\
  \And
  Zhizhong Su \\
  Senior Director \\
  Robot Lab, Horizon Robotics \\
  \texttt{zhizhongsu@horizon.auto} \\
  \And
  Jun Zhu \\
  Professor \\
  Computer Science, Tsinghua University \\
  \texttt{junzhu@mail.tsinghua.edu.cn} \\
}

\begin{document}

\maketitle

\begin{abstract}
Gradient-based adversarial attacks using the Cross-Entropy (CE) loss often suffer from overestimation due to relative errors in gradient computation induced by floating-point arithmetic. This paper provides a rigorous theoretical analysis of these errors, conducting the first comprehensive study of floating-point computation errors in gradient-based attacks across four distinct scenarios: (i) unsuccessful untargeted attacks, (ii) successful untargeted attacks, (iii) unsuccessful targeted attacks, and (iv) successful targeted attacks. We establish theoretical foundations characterizing the behavior of relative numerical errors under different attack conditions, revealing previously unknown patterns in gradient computation instability, and identify floating-point underflow and rounding as key contributors. Building on this insight, we propose the Theoretical MIFPE (\(\yuloss\)) loss function, which incorporates an optimal scaling factor \( T = t^* \) to minimize the impact of floating-point errors, thereby enhancing the accuracy of gradient computation in adversarial attacks. Extensive experiments on the MNIST, CIFAR-10, and CIFAR-100 datasets demonstrate that \(\yuloss\) outperforms existing loss functions, including CE, C\&W, DLR, and MIFPE, in terms of attack potency and robustness evaluation accuracy.
\end{abstract}

\section{Introduction}\label{sec:intro}

Deep learning has fundamentally reshaped the landscape of artificial intelligence (AI), achieving remarkable success across diverse domains. Its applications span safety-critical fields such as aviation~\cite{le2023improving}, where neural networks enable object classification in X-ray imagery, medical diagnosis~\cite{yadav2019deep}, facilitating precise segmentation of medical images, and autonomous driving~\cite{feng2020deep}, where models support real-time object detection for navigation. Moreover, the advent of large-scale language models, exemplified by ChatGPT, has further expanded the reach of deep learning, revolutionizing natural language processing and enabling applications in areas such as automated content generation, education, and human-computer interaction~\cite{tabone2023using}. As deep learning becomes increasingly integrated into human life, its applications continue to grow, permeating sectors like healthcare, education, and public policy, thereby amplifying its societal impact. 

However, these advancements are accompanied by a critical vulnerability: deep neural networks (DNNs) exhibit limited robustness to adversarial attacks. Even imperceptible perturbations to input data can mislead models into producing erroneous outputs~\cite{szegedy14,goodfellow15}, posing significant safety risks to DNN-based systems. For instance, in autonomous driving, a misclassification induced by a small perturbation could result in catastrophic failures~\cite{boloor2020attacking}. This vulnerability highlights the pressing need to enhance the robustness of DNNs and to develop reliable evaluation strategies capable of accurately assessing their resilience against adversarial threats, thereby ensuring the safe and ethical deployment of deep learning technologies in an increasingly interconnected society.

To tackle these challenges, the research community has proposed numerous defense strategies~\cite{goodfellow15,moosavi2016deepfool,carlini17,madry18,dong18momentum,xiao2019advgan,croce20aa} aimed at improving model robustness, alongside various attack techniques~\cite{madry18,shafahi19free,alayrac19,zhang19trades,pang20hypersphere,wang20misclass,wu20wp,wu20width,yu2021lafeat,gao2022mora,yu2023lafit} designed to evaluate defense effectiveness. Among attack methods, white-box attacks represent the most rigorous robustness test, as attackers possess complete knowledge of the target model's architecture, parameters, training algorithms, and datasets. A canonical example is the Projected Gradient Descent (PGD) attack~\cite{madry18}, which exploits gradient information to generate adversarial examples for resilience assessment. 

However, studies demonstrate that PGD paired with conventional cross-entropy (CE) loss frequently overestimates model robustness~\cite{croce20aa,mao2021caa,yu2021lafeat}. This occurs because CE-computed gradients inadequately guide adversarial example generation—a phenomenon associated with gradient masking~\cite{goodfellow2018gradient}. Advanced evaluation methods have thus employed attack algorithm ensembles~\cite{croce20aa,mao2021caa} to combine multiple strategies for improved accuracy. Yet these methods lack fundamental analysis of root causes behind gradient-based attack overestimation. Research by \cite{gao2022mora} reveals persistent overestimation issues even when evaluating advanced defenses like ensemble methods~\cite{kariyappa2019improving,pang2019improving,yang2020dverge}.

The failure of gradient-based attacks like PGD with CE loss has prompted the development of alternative loss functions to mitigate the overestimation problem. Notable examples include the Carlini and Wagner (C\&W) loss~\cite{carlini17}, the Difference-of-Logits Ratio (DLR) loss~\cite{croce20aa}, and the Minimize the Impact of Floating-point Errors (MIFPE) loss~\cite{yu2023efficient}. The C\&W loss, also known as the hinge loss, avoids exponential operations present in CE, reducing the risk of floating-point underflow errors, but it discards some logit elements, potentially weakening the attack's effectiveness. Similarly, the DLR loss scales the difference between the largest logits to improve gradient computation but suffers from similar limitations due to partial logit utilization.

In contrast, MIFPE~\cite{yu2023efficient}  provides a deeper insight into the root cause of overestimation. The authors reveal that the relative error in gradients, induced by floating-point errors—such as underflow and rounding errors—is a fundamental reason for the failure of CE-based gradient attacks. This relative error is particularly influenced by the difference between the first and second largest logits (\(\Delta = \mathbf{z}_{\pi_1} - \mathbf{z}_{\pi_2} \)). To mitigate this, MIFPE scales the logits by a factor \(c=T/\Delta_{detach}\) where \( T = 1 \), significantly reducing the impact of floating-point errors and improving the accuracy of robustness evaluation. Experimental results demonstrate that MIFPE outperforms CE, C\&W, and DLR in terms of efficiency and reliability across various defense mechanisms. However, a critical limitation of MIFPE lies in its empirical determination of \( T = 1 \), which lacks a theoretical foundation, raising questions about its optimality.

In this paper, we extend the foundational work of MIFPE~\cite{yu2023efficient} by pioneering a comprehensive theoretical framework that dissects relative errors in gradients caused by floating-point arithmetic in the context of the cross-entropy (CE) loss function. Our groundbreaking analysis is the first to systematically explore floating-point computation errors across four distinct attack scenarios: (i) unsuccessful untargeted attacks, (ii) successful untargeted attacks, (iii) unsuccessful targeted attacks, and (iv) successful targeted attacks. By establishing robust theoretical foundations, we uncover novel insights into the patterns of numerical error behavior under varying attack conditions, shedding light on previously unrecognized instabilities in gradient computations.
Building on this theoretical insight, we propose the Theory Version of  Minimize the Impact of Floating Point Error  \((\yuloss)\) loss function, which adopts the optimal scaling factor \( T = t^* \). This adjustment ensures that the relative error due to floating-point errors is minimized, thereby enhancing the accuracy of gradient computation in adversarial attacks. Experimental results on the MNIST, CIFAR-10, and CIFAR-100 datasets validate that the \(\yuloss\) loss function outperforms existing loss functions, including CE, C\&W, DLR, and MIFPE, in terms of attack potency and robustness evaluation accuracy. To summarize, our main contributions are as follows:

\begin{itemize}
    \item We conduct the first comprehensive study of floating-point computation errors in gradient-based attacks across four distinct scenarios: (i) unsuccessful untargeted attacks, (ii) successful untargeted attacks, (iii) unsuccessful targeted attacks, and (iv) successful targeted attacks.
    \item We establish theoretical foundations characterizing the behavior of relative numerical errors under different attack conditions, revealing previously unknown patterns in gradient computation instability.
    \item Leveraging this theoretical analysis, we propose the \(\yuloss\) loss function, which incorporates an optimal scaling factor \( T = t^* \) to minimize the impact of floating-point errors on gradient computation.
    \item Through extensive experiments, we demonstrate that the \(\yuloss\) loss function outperforms existing methods, including CE, C\&W, DLR, and MIFPE, in terms of attack potency and robustness evaluation accuracy.
\end{itemize}

\section{%
    Preliminaries \& Related Work%
}\label{sec:related}

\subsection{Preliminaries}

To lay the groundwork for our study, we formally introduce the key concepts of adversarial examples and white-box attacks in the context of deep neural networks (DNNs). Consider a DNN classifier \( f_{\boldsymbol{\theta}} : \mathbb{R}^{C \times H \times W} \to \mathbb{R}^K \), parameterized by \( \boldsymbol{\theta} \), which maps an input image \( \mathbf{x} \in \mathbb{R}^{C \times H \times W} \) to output logits \( \mathbf{z} \in \mathbb{R}^K \). Here, \( C \), \( H \), and \( W \) denote the number of channels, height, and width of the input, respectively, while \( K \) represents the number of classes. The true label of \( \mathbf{x} \) is denoted by \( y \), and the predicted class is determined as \( \arg\max f_{\boldsymbol{\theta}}(\mathbf{x}) \).

Adversarial attacks seek to craft adversarial examples \( \mathbf{\hat{x}} \) that cause the model to misclassify, i.e., \( \arg\max f_{\boldsymbol{\theta}}(\mathbf{\hat{x}})_i \neq y \). This is achieved by introducing a perturbation \( \mathbf{\delta} = \mathbf{\hat{x}} - \mathbf{x} \), constrained by \( \|\mathbf{\delta}\|_p \leq \epsilon \), where \( \epsilon \) is the perturbation magnitude and \( p \) specifies the norm (e.g., \( \ell_\infty \), \( \ell_2 \)). The adversarial example is generated by solving the optimization problem:
\begin{equation}
\mathbf{\hat{x}} = \mathbf{x} + \arg\max_{\|\mathbf{\delta}\|_p \leq \epsilon} L(f_{\boldsymbol{\theta}}(\mathbf{x} + \mathbf{\delta}), y),
\label{eq:adversarial}
\end{equation}
where \( L \) is the loss function, typically cross-entropy (CE) for classification tasks.

White-box attacks assume full knowledge of the model’s architecture, parameters, and training data, posing the most stringent challenge for defense mechanisms. Gradient-based methods, such as Projected Gradient Descent (PGD)~\cite{madry18}, are widely employed to solve~\eqref{eq:adversarial}. PGD iteratively updates the adversarial example via:

\begin{equation}
\mathbf{\hat{x}}_{i+1} = \text{Proj}_{\mathcal{B}_\epsilon(\mathbf{x})} \left( \mathbf{\hat{x}}_{i} + \alpha_i \cdot \text{sign} \left( \nabla_{\mathbf{\hat{x}}_{i}} L(f_{\boldsymbol{\theta}}(\mathbf{\hat{x}}_{i}), y) \right) \right),
\label{eq:pgd}
\end{equation}
where $\mathbf{\hat{x}}_0 = \text{Proj}_{\mathcal{B}_\epsilon(\mathbf{x})} (\mathbf{x} + \mathbf{u})$ is the initial adversarial example, with $\mathbf{u} \sim \text{Uniform}[-\epsilon, \epsilon]$ being a random perturbation.
 $\alpha_i$ is the step size at iteration $i$.
 $\text{Proj}_{\mathcal{B}_\epsilon(\mathbf{x})}$ denotes the projection operator that ensures the adversarial example remains within the $\epsilon$-ball centered at $\mathbf{x}$ under the $\ell_p$-norm (typically $\ell_\infty$ or $\ell_2$).
 $\nabla_{\mathbf{\hat{x}}_{i}} L$ is the gradient of the loss function $L$ with respect to the adversarial example $\mathbf{\hat{x}}_{i}$.

The effectiveness of PGD relies on the choice of the loss function $L$,  a central focus of our investigation.

\subsection{Related Work}

The susceptibility of deep neural networks (DNNs) to adversarial attacks has spurred extensive research into both attack methodologies and defense strategies, with the primary objectives of evaluating model robustness and enhancing resilience against adversarial perturbations. Adversarial attacks are categorized into white-box and black-box settings, where white-box attacks exploit complete knowledge of the model’s architecture, parameters, and training data to craft targeted perturbations. Traditional white-box attack methods, such as the Fast Gradient Sign Method (FGSM)~\cite{goodfellow15}, Basic Iterative Method (BIM)~\cite{kurakin17adversarial}, Momentum Iterative Method (MIM)~\cite{dong18momentum}, Projected Gradient Descent (PGD)~\cite{madry18}, and Fast Adaptive Boundary Attack (FAB)~\cite{croce20fab}, predominantly target \( \ell_\infty \)\-norm perturbations. Additionally, methods like Carlini and Wagner (C\&W)~\cite{carlini17} and DeepFool~\cite{moosavi2016deepfool} are designed for \( \ell_2 \)\-norm attacks. These approaches have been widely adopted to assess adversarial robustness; however, extensive empirical evidence has revealed their significant limitation in overestimating model robustness. 

To address the pervasive issue of overestimating model robustness, researchers have proposed strategies that integrate multiple attack methods, as relying on any single attack method often fails to provide an accurate assessment of robustness. A prominent example is AutoAttack~\cite{croce20aa}, an ensemble-based approach that combines both white-box and black-box attack strategies. AutoAttack has become the de facto standard for benchmarking adversarial robustness due to its comprehensive evaluation capabilities. However, recent studies have demonstrated that superior attack performance can be achieved without integrating multiple attack methods. A notable example is LAFEAT~\cite{yu2021lafeat}, which leverages latent feature representations to enhance attack efficacy. While these strategies have significantly alleviated the overestimation of model robustness, they often incur substantial computational overhead. This high computational cost limits their applicability to large-scale or real-time scenarios. Moreover, these methods lack a fundamental analysis of the root causes behind the overestimation of robustness in gradient-based attacks. Research by~\cite{gao2022mora} has shown that these approaches still exhibit significant overestimation issues when evaluating advanced defense strategies, such as ensemble defenses~\cite{kariyappa2019improving, pang2019improving, yang2020dverge}.

In an effort to uncover the root causes of robustness overestimation in gradient-based attacks, Yu et al.~\cite{yu2023efficient} identified that floating-point arithmetic errors introduce relative errors in the computed gradients, which contribute to the overestimation problem. To address this, they proposed a novel loss function, MIFPE (Minimizing Floating-Point Error), designed to mitigate the negative impact of floating-point errors on gradient-based attacks. The MIFPE loss function is defined as:

\begin{equation}
    \mathcal{L}^{\text{MIFPE}}\left( \mathbf{z}, y \right) \triangleq \mathcal{L}^{\text{CE}}\left( T \cdot \mathbf{z} / \Delta_{\text{value}}, y \right),
\end{equation}
where \( T = 1 \) is the scale factor, \( \Delta = \mathbf{z}_{\pi 1} - \mathbf{z}_{\pi 2} \), and \(\text{value}\) is an operation that removes gradient information from \(\Delta_{\text{value}}\), distinguishing it from \(\Delta\), which retains gradient information.

Despite its effectiveness, the empirical choice of \( T = 1 \) in MIFPE lacks theoretical justification, raising concerns about its optimality. This limitation suggests that the relative error in gradients may not be fully minimized, potentially leading to suboptimal attack performance.

\section{Theory Analysis}\label{sec:method_untargeted}

Assuming the input \( \mathbf{\hat{x}} \) is correctly labeled with \( y \), we compute the model’s output logits as \( \mathbf{z} = f_{\boldsymbol{\theta}}(\mathbf{\hat{x}}) \), where \( \boldsymbol{\theta} \) denotes the model parameters. Subsequently, we sort the elements of \( \mathbf{z} \) in descending order, with \( \mathbf{z}_{\pi_1} \) representing the maximum value, typically corresponding to the predicted class under normal conditions. This setup provides the foundation for analyzing adversarial attacks, which perturb \( \mathbf{\hat{x}} \) to exploit vulnerabilities in the model’s decision-making process.

Adversarial attacks are broadly classified into untargeted and targeted variants, distinguished by their objectives and loss formulations. An untargeted attack employs the cross-entropy loss \( CE(\mathbf{z}, y) \) to maximize the deviation of the model’s prediction from the true label \( y \), inducing misclassification into any incorrect class. In contrast, a targeted attack leverages the negative cross-entropy loss \( -CE(\mathbf{z}, y_t) \), where \( y_t \) is the attacker-specified target label, aiming to steer the prediction precisely toward \( y_t \). This distinction necessitates separate analyses of the relative gradient errors---arising from floating-point arithmetic inaccuracies---when \( CE \) serves as the loss function, as the attack type influences the gradient computation.

To systematically analyze the relative errors in gradient computations across different attack scenarios, we formally define four distinct error metrics corresponding to each operational phase: (i) $\delta_{u\text{-}u}$ for untargeted attacks in unsuccessful phases, (ii) $\delta_{u\text{-}s}$ for untargeted attacks in successful phases, (iii) $\delta_{t\text{-}u}$ for targeted attacks during unsuccessful attempts, and (iv) $\delta_{t\text{-}s}$ for successful targeted attacks, where these notations will be consistently employed throughout our subsequent analysis.

\subsection{Relative  Error of Gradient in Untargeted Attacks}\label{subsec:untargeted}

First, we examine the relative error in the computed gradients due to floating-point inaccuracies under untargeted attacks.
In untargeted adversarial attacks, the attacker's primary goal is to maximize the value of \( \max_{i \neq y} \mathbf{z}_i  -\mathbf{z}_y \), transforming it from a negative value (indicating correct classification) to a positive value (indicating misclassification).

\subsubsection{Unsuccessful Attack Phase}\label{sec:method_untargeted_unsuccess}
{ When \(\mathbf{z}_y = \mathbf{z}_{\pi_1}\)}

\begin{equation}
CE\left( \mathbf{z,}y \right) =-\log p_y=-\log \frac{e^{\mathbf{z}_y-\mathbf{z}_{\pi 1}}}{\sum_{i=1}^K{e^{\mathbf{z}_i-\mathbf{z}_{\pi 1}}}}
\label{eq:SCE_u}
\end{equation}

\begin{equation}
CE\left( c\mathbf{z},y \right) =-\log p_y^{c}=-\log \frac{e^{c(\mathbf{z}_y-\mathbf{z}_{\pi_1})}}{\sum_{i=1}^K{e^{c(\mathbf{z}_i-\mathbf{z}_{\pi_1})}}}
\label{eq:SCE_c_u}
\end{equation}

\begin{equation}
\begin{split}
 \nabla _{\mathbf{\hat{z}}}CE\left(  \text{c}\mathbf{z,}y \right)   & = c\left( -1+p_{y}^{c} \right) \nabla _{\mathbf{\hat{x}}}\left( \mathbf{z}_y-\mathbf{z}_{\pi 1} \right)   \quad+\sum_{i\ne y}{cp_{i}^{c}\nabla _{\mathbf{\hat{x}}}\left( \mathbf{z}_i-\mathbf{z}_{\pi 1} \right)}\\
 &=c\sum_{i\ne {\pi_{1}}}{p_{i}^{c}\nabla _{\mathbf{\hat{x}}}\left( \mathbf{z}_i-\mathbf{z}_{\pi_{1}} \right)} \\
 &= c  p_{\pi_{2}}^{c} \nabla _{\mathbf{\hat{x}}} \left(  \mathbf{z}_{\pi_{2}} -  \mathbf{z}_{\pi_{1}} \right)  \\
 &\quad + c  p_{\pi_{3}}^{c} \nabla _{\mathbf{\hat{x}}} \left( \mathbf{z}_{\pi_{3}} -\mathbf{z}_{\pi_{2}}+  \mathbf{z}_{\pi_{2}} -  \mathbf{z}_{\pi_{1}} \right) \\
 & \quad + ...\\
 &\quad + c  p_{\pi_{K}}^{c} \nabla _{\mathbf{\hat{x}}} \left( \mathbf{z}_{\pi_{K}} -\mathbf{z}_{\pi_{K-1}}+ ...+ \mathbf{z}_{\pi_{2}} -  \mathbf{z}_{\pi_{1}} \right) \\
&=c(1-p_{\pi_{1}}^{c})\nabla _{\mathbf{\hat{x}}}\left( \mathbf{z}_{\pi_{2}}-\mathbf{z}_{\pi_{1}} \right)  \\
&\quad + c(1-p_{\pi_{1}}^{c}-p_{\pi_{2}}^{c})\nabla _{\mathbf{\hat{x}}}\left( \mathbf{z}_{\pi_{3}}-\mathbf{z}_{\pi_{2}} \right)  \\
& \quad + ... \\
&\quad + c(1-p_{\pi_{1}}^{c}-...-p_{\pi_{K-1}}^{c})\nabla _{\mathbf{\hat{x}}}\left( \mathbf{z}_{\pi_{K}}-\mathbf{z}_{\pi_{K-1}} \right)  \\
\end{split}
\end{equation}

where \(c=\frac{t}{\Delta_{\text{value}} }\) is a scale factor, \(t>0\), and $p_{i}^{c}=e^{c\left( \mathbf{z}_i-\mathbf{z}_{\pi 1} \right)}/\sum_{j=1}^K{e^{c\left( \mathbf{z}_j-\mathbf{z}_{\pi 1} \right)}}$.

The notation $\Delta_{\text{value}}$ explicitly distinguishes the scalar magnitude from its differentiable counterpart $\Delta = \mathbf{z}_{\pi_1} - \mathbf{z}_{\pi_2}$. Since $\Delta$ inherits gradient information through its dependence on the logits $\mathbf{z}$, we employ the detachment operation $\Delta_{\text{value}} = \Delta.\text{detach}()$ to isolate the pure numerical value. This ensures that the scaling factor $c = t/\Delta_{\text{value}}$ remains a gradient-free constant during computation.

\begin{equation}
 \max_{i \neq y} \mathbf{z}_i -\mathbf{z}_y   =   \mathbf{z}_{\pi_2} -\mathbf{z}_{\pi_1} 
 \label{eq:untargeted_obj_unsuccess}
\end{equation}

As derived from the primary objective of untargeted attacks in Equation~\eqref{eq:untargeted_obj_unsuccess}, the gradient term $\nabla_{\mathbf{\hat{x}}}(\mathbf{z}_{\pi_2} - \mathbf{z}_{\pi_1})$ emerges as a critical component. During the unsuccessful attack phase, we consequently focus on minimizing the relative error in $|c(1-p_{\pi_1}^c)\nabla_{\mathbf{\hat{x}}}(\mathbf{z}_{\pi_2}-\mathbf{z}_{\pi_1})|$.

\begin{equation}
\begin{split}
\delta{( t)}_{u\_u}  = \frac{\epsilon}{ |c (1 - p_{\pi_1}^c) \nabla _{\mathbf{\hat{x}}}\left( \mathbf{z}_{\pi_2}-\mathbf{z}_{\pi_1} \right)|}
\end{split}
\end{equation}

where \(\epsilon\) denotes the error incurred due to floating-point truncation in the numerical computation of \( |c (1 - p_{\pi_1}^c) \nabla _{\mathbf{\hat{x}}}\left( \mathbf{z}_{\pi_2}-\mathbf{z}_{\pi_1} \right)| \), with \(\epsilon\) constrained within the range \( 0 \leq \epsilon \leq \epsilon_{\text{max}} \), where 
\(\epsilon_{\text{max}}\) is the upper bound of \(\epsilon\), which is determined by the precision of the floating-point format, corresponding to \( 2^{-10} \) for 16-bit floating-point numbers, \( 2^{-23} \) for 32-bit floating-point numbers, and \( 2^{-52} \) for 64-bit floating-point numbers.

In our analysis, we establish that \(\delta(t) = 1\) when \( |c (1 - p_{\pi_1}^c) \nabla _{\mathbf{\hat{x}}}\left( \mathbf{z}_{\pi_2}-\mathbf{z}_{\pi_1} \right)| \) encounters floating-point underflow, as the truncation error \(\epsilon\) then equals \( |c (1 - p_{\pi_1}^c) \nabla _{\mathbf{\hat{x}}}\left( \mathbf{z}_{\pi_2}-\mathbf{z}_{\pi_1} \right)| \). However, in cases where this term neither underflows nor overflows, the exact value of \(\epsilon\) becomes contingent upon the specific numerical truncation induced by floating-point operations, introducing variability that complicates precise error estimation. To address this uncertainty and provide a reliable basis for evaluation, we focus on the deterministic upper bound of the truncation error  \(\epsilon_{\text{max}}\), which is both well-defined and independent of implementation-specific truncation effects. Analyzing the worst-case scenario—where \(\epsilon = \epsilon_{\text{max}}\)—is critical to rigorously assess the maximum impact of floating-point truncation on relative error. This conservative approach enables us to derive an exact upper bound, \(\delta{( t)}_{u\_u}^{\text{sup}}\), for the relative error in gradient computations, ensuring a robust and reproducible framework for understanding the limitations imposed by floating-point arithmetic in adversarial attack strategies.

\begin{equation}
\begin{split}
\delta{( t)}_{u\_u}^{\text{sup}}
= \sup_{\epsilon \in [0, \epsilon_{\text{max}}]} \delta{( t,\epsilon)}_{u\_u}  =\frac{\epsilon_{\text{max}}}{ |c (1 - p_y^c) \nabla _{\mathbf{\hat{x}}}\left(  \mathbf{z}_{\pi_2} -\mathbf{z}_{\pi_1} \right)|}
\end{split}
\end{equation}

Consequently, the minimum value of \(\delta{( t)}_{u\_u}^{\text{sup}}\) can be expressed as:

\begin{equation}
\begin{split}
 \delta{( t)}_{u\_u}^{\text{sup}\_\text{min}} 
 = \frac{\epsilon_{\text{max}}}{  {|c (1 - p_{\pi_1}^c) \nabla _{\mathbf{\hat{x}}}\left(  \mathbf{z}_{\pi_2} -\mathbf{z}_{\pi_1} \right)|}_{max} },
\end{split}
\end{equation}

where \({ |c (1 - p_{\pi_1}^c) \nabla _{\mathbf{\hat{x}}}\left( \max_{i \neq {\pi_1}} \mathbf{z}_i -\mathbf{z}_{\pi_1} \right)| }_{max}\) denotes the maximum value of the denominator across the relevant domain.

Gradient-based iterative attacks involve repeatedly applying a uniform procedure at each iteration, where perturbations are introduced to the input data based on gradient information derived through backpropagation. Given the repetitive nature of this mechanism, the overall multi-iteration process can be effectively understood by analyzing a single iteration in detail. Consequently, we focus our analysis on the relative error incurred during the gradient computation phase via backpropagation within a specific iteration of the multi-iteration attack. Notably, the gradient \( \nabla _{\mathbf{\hat{x}}}\left( \mathbf{z}_{\pi_2}-\mathbf{z}_{\pi_1} \right) \), computed in this process, depends solely on the model’s internal parameters and remains invariant to the scaling factor \( t / \Delta_{\text{value}} \) incorporated into the loss function. Thus, we treat \( \nabla _{\mathbf{\hat{x}}}\left( \mathbf{z}_{\pi_2}-\mathbf{z}_{\pi_1}\right) \) as a constant with respect to \( t / \Delta_{\text{value}} \). As a result, maximizing \( |c (1 - p_{\pi_1}^c) \nabla _{\mathbf{\hat{x}}}\left( \mathbf{z}_{\pi_2}-\mathbf{z}_{\pi_1} \right)| \) simplifies to maximizing \(  c (1 - p_{\pi_1}^c) \), since the gradient term is constant. To analyze the maximum value of \(   c (1 - p_{\pi_1}^c)   \), we define a new function \( {g(t)_{u\_u}} \) as follows:

\begin{equation}
{g( t)}_{u\_u} = c \left( 1 - p_{\pi_1}^c \right)>0
\end{equation}

\begin{equation}
{g'(t)}_{u\_u} = \frac{B^2 - B + c S}{\Delta_{\text{value}} B^2}.
\end{equation}
where $p_{\pi_1}^c = \frac{e^{c \cdot 0}}{\sum_{j=1}^K e^{c (\mathbf{z}_j - \mathbf{z}_{\pi_1})}} = \frac{1}{B}, \quad B = \sum_{j=1}^K e^{c (\mathbf{z}_j - \mathbf{z}_{\pi_1})} = 1 + \sum_{j \neq {\pi_1}} e^{c (\mathbf{z}_j - \mathbf{z}_{\pi_1})}> 1$
, and $ S = \sum_{j=1}^K (\mathbf{z}_j - \mathbf{z}_{\pi_1}) e^{c (\mathbf{z}_j - \mathbf{z}_{\pi_1})}<0 $ , due to \(\mathbf{z}_j - \mathbf{z}_{\pi_1}<0\), \( c = t / \Delta_{\text{value}} \), and \( \Delta_{\text{value}} = \mathbf{z}_{\pi_1} - \mathbf{z}_{\pi_2}>0\).

To investigate the behavior of \( g'(t) \), we define an auxiliary function as follows:
\begin{equation}
h(t) = B^2 - B + c S 
\end{equation}
The function \( h(t) \) is linear with a slope of \( c=S / \Delta_{\text{value}}<0 \), indicating that it is monotonically decreasing.
We define \( t^* \) as the point where \( g'(t^*) = 0 \), given by:
\begin{equation}
t^* = \frac{\Delta_{\text{value}} B (B - 1)}{-S},
\label{eq:t_star}
\end{equation}
Where \( \Delta_{\text{value}} > 0 \), \( B > 1 \), and \( S < 0 \), ensuring \( t^* > 0 \). At \( t = t^* \), \( g(t)_{u\_u} \) achieves its maximum value.

\subsubsection{Successful Attack Phase}
Due to space limitations, we defer the detailed theoretical analysis of the relative gradient error \(\delta_{u\text{-}s}\) for the successful phase of untargeted attacks to Appendix~\ref{sec:method_untargeted_success}.

\subsection{Relative  Error of Gradient in Targeted Attacks}

Next, we analyze the relative error in the computed gradients due to floating-point inaccuracies under targeted attacks. 
In targeted adversarial attacks, the attacker's primary goal is to maximize the value of  \( \mathbf{z}_{y_t} - \max_{i \neq y_t} \mathbf{z}_i \), transforming it from a negative value (indicating unsuccessful targeted attacks) to a positive value (signifying successful target misclassification)

Comprehensive theoretical analyses of the relative gradient errors \(\delta_{t\text{-}u}\) for unsuccessful targeted attacks and \(\delta_{t\text{-}s}\) for successful targeted attacks are provided in Appendix~\ref{sec:method_targeted_unsuccess} and Appendix~\ref{sec:method_targeted_success}, respectively.

We visualize the numerical behavior of \( g(t) \) and \( \delta_{\text{sup}}(t) \) corresponding to four distinct scenarios: (1) untargeted attacks in unsuccessful phases, (2) untargeted attacks in successful phases, (3) targeted attacks in unsuccessful phases, and (4) targeted attacks in successful phases in Figure~\ref{fig:targeted_and_untargeted}. As established by our prior analysis and illustrated in Figure~\ref{fig:targeted_and_untargeted}, the optimal value \( t^* \) is contingent upon the model’s output logits \( \mathbf{z} \) and the specific scenario among the four outlined cases. In the context of multi-round gradient-based attacks, the logits \( \mathbf{z} \) evolve with each iteration due to perturbations applied to the input data, resulting in a dynamic process where the model may correctly classify certain inputs while misclassifying others. Consequently, employing a fixed \( t = 1 \) within the MIFPE framework is suboptimal, as it fails to adapt to these iterative changes. To address this limitation, we propose recomputing \( t^* \) before each attack iteration based on the updated \( \mathbf{z} \) and the prevailing scenario. Leveraging this adaptive approach, we introduce the theory version of  Minimize the Impact of Floating Point Error loss function, denoted as \( \yuloss \), defined as follows:

\begin{figure*}[htbp]
    \begin{subfigure}[b]{0.48\textwidth}
        \centering
        \includegraphics[width=\linewidth]{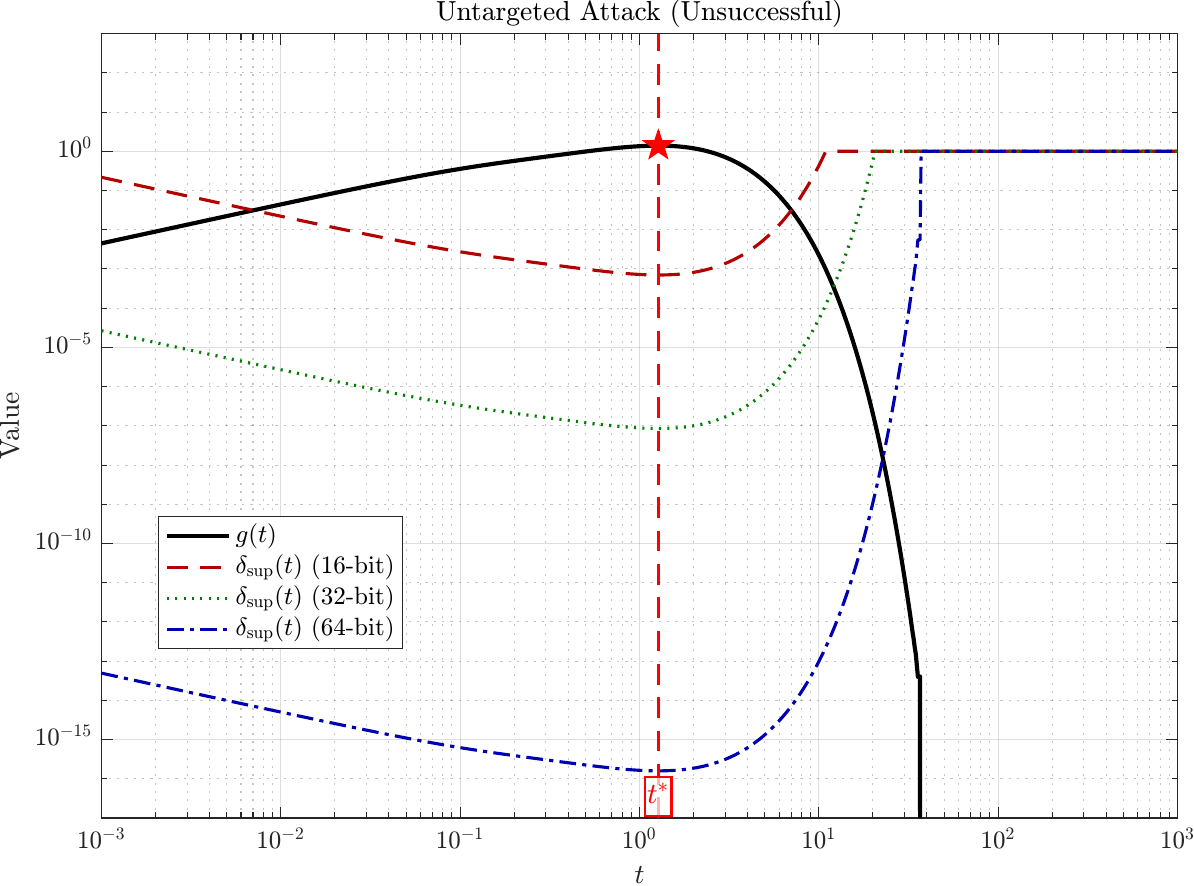}
        \caption{Untargeted Attack Unsuccessful   Phase   }
        \label{fig:a}
    \end{subfigure}
    \hfill
    \begin{subfigure}[b]{0.48\textwidth}
        \centering
        \includegraphics[width=\linewidth]{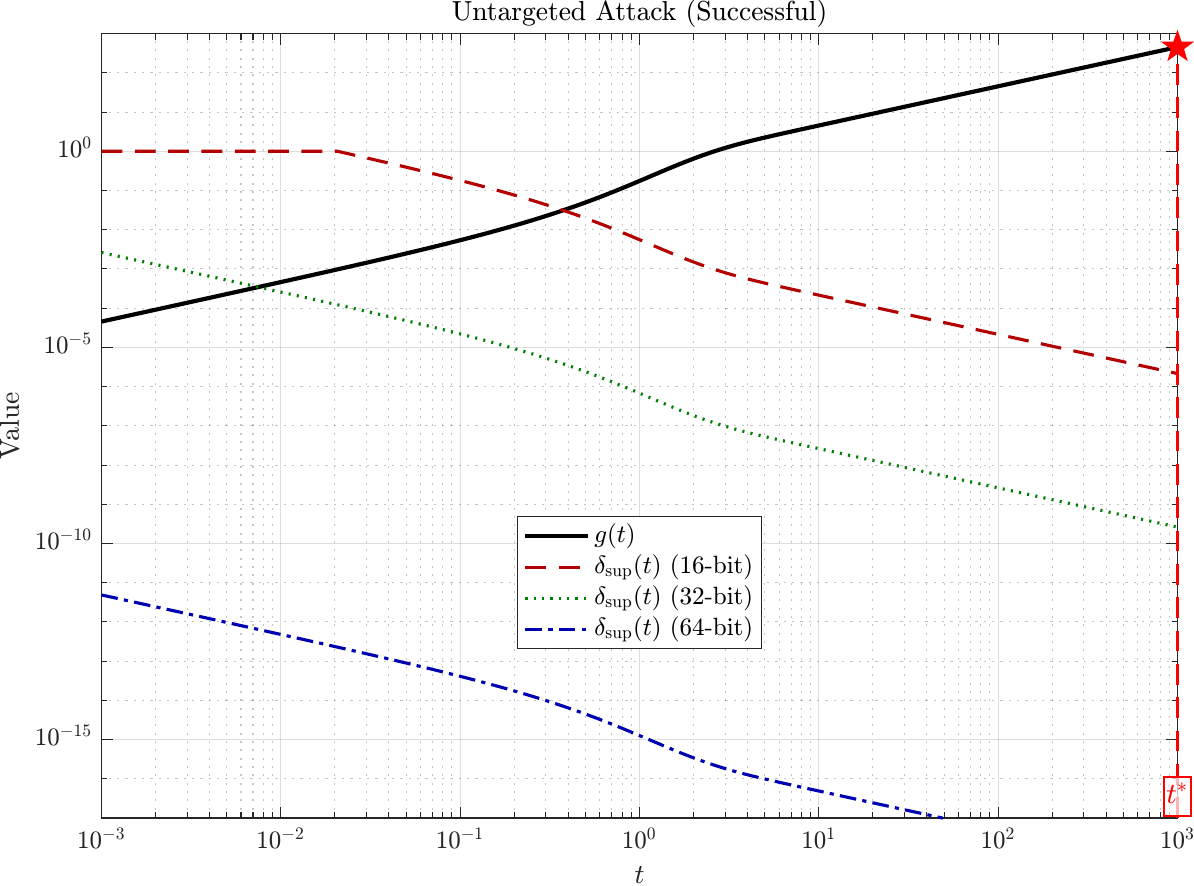}
        \caption{Untargeted Attack Successful  Phase   }
        \label{fig:b}
    \end{subfigure}
    
    \begin{subfigure}[b]{0.48\textwidth}
        \centering
        \includegraphics[width=\linewidth]{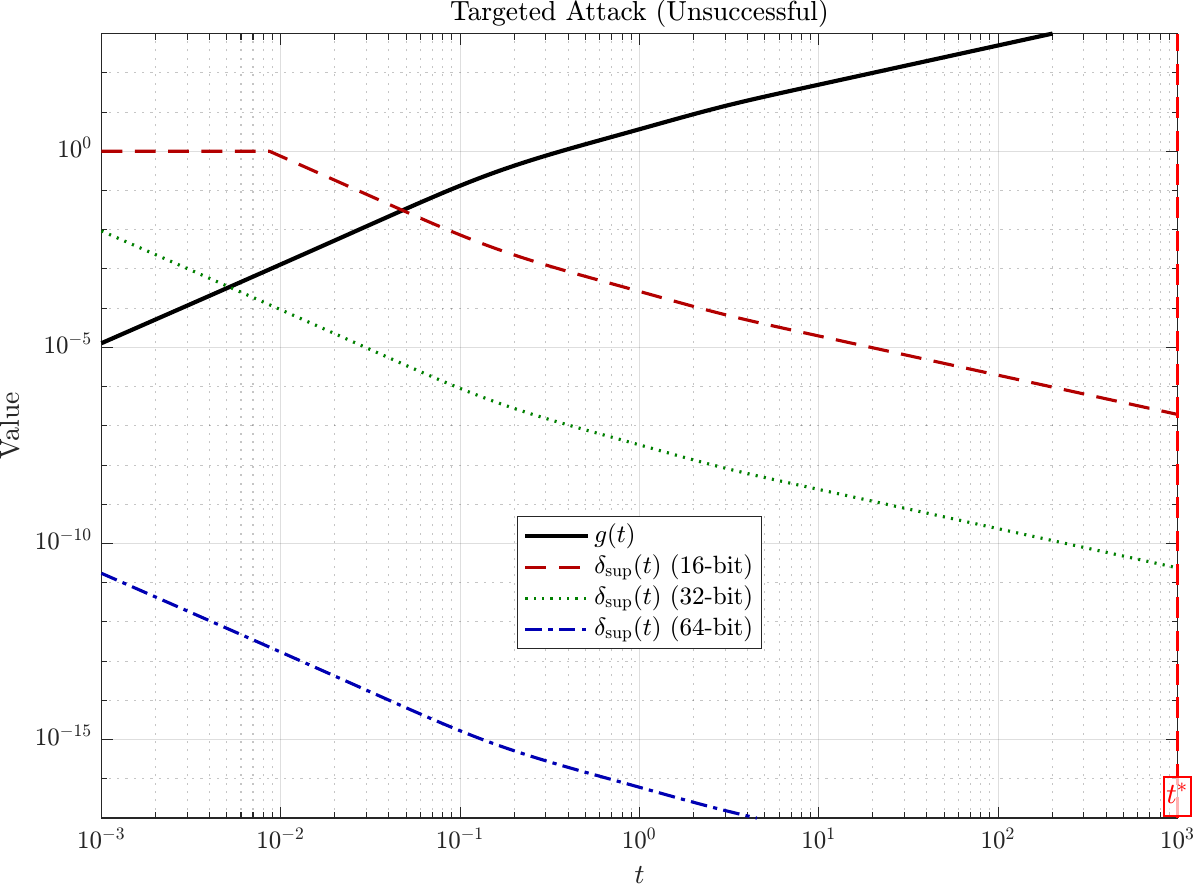}
        \caption{Targeted Attack Unsuccessful  Phase   }
        \label{fig:c}
    \end{subfigure}
    \hfill
    \begin{subfigure}[b]{0.48\textwidth}
        \centering
        \includegraphics[width=\linewidth]{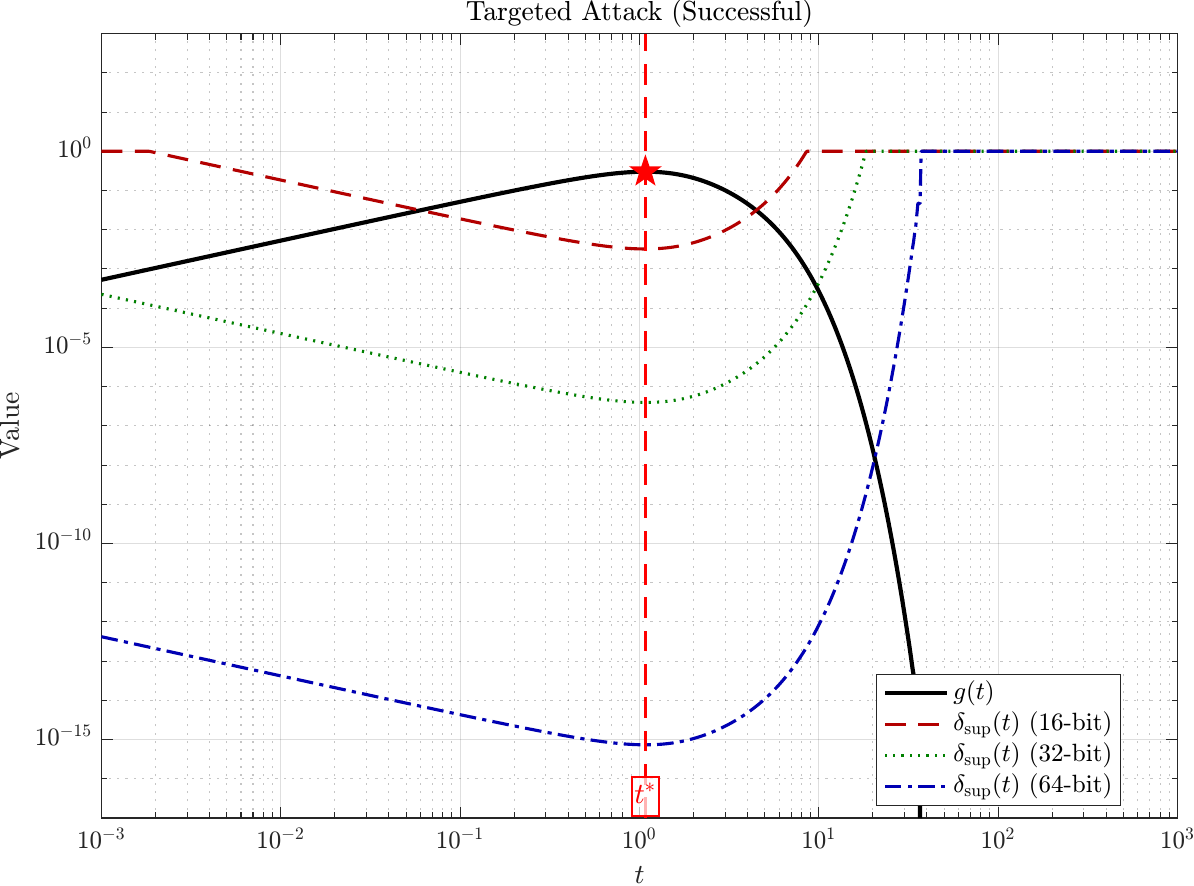}
        \caption{Targeted Attack Successful   Phase      }
        \label{fig:d}
    \end{subfigure}
    
    \caption{
        Analysis of the relative error in gradients computed using cross-entropy loss for a 10-class classifier under targeted and untargeted attack scenarios.  
\textbf{Untargeted Attack Scenario} :  
(a) \textit{Untargeted Attack Unsuccessful Attack Phase (\( \mathbf{z}_y = \mathbf{z}_{\pi_1} \))}: (\( g(t) = c (1 - p_{\pi_1}^c) \)) Logits \( \mathbf{z} = [2.5, -1.3, 0.8, 3.8, -0.9, 1.7, -2.1, 3.6, 0.4, -1.5] \), with \( y = 3 \) as the ground-truth label and \( \mathbf{z}_{\pi_1} = 3.8 \).  
(b) \textit{Untargeted Attack Successful Attack Phase (\( \mathbf{z}_y \neq \mathbf{z}_{\pi_1} \))}: (\( g(t) = c  p_{\pi_1}^c  \)) Logits \( \mathbf{z} = [1.2, -1.5, 0.5, 1.9, -1.2, 4.2, -2.3, 2.0, 0.1, -1.8] \), with \( y = 5 \) as the misclassification label and \( \mathbf{z}_{\pi_1} = 4.2 \).  
\textbf{Targeted Attack Scenario} :  
(c) \textit{Targeted Attack Unsuccessful Attack Phase  (\( \mathbf{z}_{y_t} \neq \mathbf{z}_{\pi_1} \))}: (\( g(t) = c ( p_{\pi_1}^c - p_{y_t}^c)  \))Original logits \( \mathbf{z} = [2.5, -1.3, 0.8, 3.8, -0.9, 1.7, -2.1, 3.6, 0.4, -1.5] \), with \( y = 3 \) (correct) and target label \( y_t \neq 3 \), where \( \mathbf{z}_{\pi_1} = 3.8 \). 
(d) \textit{Targeted Attack Successful Attack Phase (\( \mathbf{z}_{y_t} = \mathbf{z}_{\pi_1} \))}: (\( g(t) = c (1 - p_{\pi_1}^c) \)) Perturbed logits \( \mathbf{z} = [1.0, 4.5, 0.3, 1.2, -1.0, 1.5, -2.5, 2.8, 0.0, -1.8] \), achieving targeted misclassification to \( y_t = 1 \), with \( \mathbf{z}_{\pi_1} = 4.5 \).  
Each subplot illustrates: (1) Upper bounds of the relative error \( \delta_{\text{sup}}(t) \) for 16-bit (red dashed), 32-bit (blue dotted), and 64-bit (green dash-dotted) floating-point precision; (2) Critical points \( t^* \) (red stars) marking the maxima of \( g(t) \), where \( \delta_{\text{sup}}(t) \) reaches local minima (gray vertical dashed lines).
        \label{fig:targeted_and_untargeted}
    }
\end{figure*}

\begin{equation}
\mathcal{L}^{\text{\yuloss}}\left( \mathbf{z}, y \right) \triangleq \mathcal{L}^{\text{ce}}\left( \frac{t^* \mathbf{z}}{(\mathbf{z}_{\pi_1}- \mathbf{z}_{\pi_2})_{\text{value}}}, y \right)
\end{equation}

\begin{equation}
\mathcal{L}_{\text{target}}^{\text{\yuloss}}\left( \mathbf{z}, y_{\text{t}} \right) \triangleq -\mathcal{L}^{\text{ce}}\left( \frac{t^* \mathbf{z}}{(\mathbf{z}_{\pi_1}- \mathbf{z}_{\pi_2})_{\text{value}}}, y_{\text{t}} \right)
\end{equation}

\section{%
Experiments%
}\label{sec:results}

To rigorously assess the effectiveness of our newly proposed \(\yuloss\) loss function in comparison to its predecessor and other established methods, we conducted a series of experiments across multiple threat models and benchmark datasets. Specifically, we evaluated the performance under \( \ell_\infty \)-bounded perturbations on the MNIST~\cite{lecun1998gradient}, CIFAR-10, and CIFAR-100~\cite{krizhevsky2010convolutional} datasets, as well as the \( \ell_2 \)-bounded threat model on the CIFAR-10 dataset. In these experiments, we utilized the PGD attack framework, incorporating a range of loss functions: the cross-entropy (CE) loss, the C\&W loss~\cite{carlini17}, the DLR loss~\cite{croce20aa}, the MIFPE loss~\cite{yu2023efficient},  and the newly introduced \(\yuloss\) loss. To ensure reproducibility and mitigate the effects of randomness, we fixed the random seed equal to zero across all trials. The experimental configuration was standardized, with each attack executed over 100 iterations using a cosine decaying step-size schedule defined as \( \epsilon_i = \epsilon (1 + cos(\pi i/I)) \), where \( I = 100 \) denotes the total iterations and \( i \) represents the current iteration. Momentum-based updates~\cite{croce20aa} with a momentum factor of 0.75 were applied, and we preserved the most potent adversarial examples identified throughout the attack process. To further benchmark the robustness of the \(\yuloss\) loss under constrained iterations, we compared its performance to the highest robust accuracy figures from RobustBench~\cite{croce2020robustbench}, which leverages an ensemble of attacks with at least 4900 iterations. Additionally, we examined the attack strength of the \(\yuloss\) loss against GAMA-PGD~\cite{sriramanan2020guided} within the same 100-iteration limit, providing a comprehensive evaluation of its efficacy.

\begin{table*}[ht]
\centering%
\caption{%
    Comparing the proposed \yuloss\ loss (\( \lnscelossyu \)), against CE (\( \sceloss \)), C\&W (\( \cwloss \)),  DLR (\( \dlrloss \)) and \yyrloss (\( \lnsceloss \)) losses. For each surrogate loss, we use PGD-100 with the step-size schedule \( \epsilon_i = \epsilon (1 + cos(\pi i/I)) \), where \( I = 100 \) denotes the total iterations and \( i \) represents the current iteration.  and momentum \( \momentum = 0.75 \) . 
    Numbers in parentheses indicate the improvement 
    w.r.t\
    the CE baseline.
    The AA, calculated using an ensemble of attacks and a minimum of 4900 iterations, is reported from RobustBench  to demonstrate how closely \yuloss\ can approach the lowest known robustness accuracy with only 100 iterations.
}\label{tab:loss}
\adjustbox{width=\textwidth}{%
\begin{tabular}{l|c|L|L|Ll|Ll|Ll|Ll|Ll|L}
\toprule
\textbf{Defense method}
& \multicolumn{1}{|c|}{\textbf{Architecture}}
& \multicolumn{1}{|c|}{\textbf{Clean}}
    & \thead{CE (\( \sceloss \))}
    & \multicolumn{2}{|c}{\textbf{C\&W (\( \cwloss \)}) }
    & \multicolumn{2}{|c}{\textbf{DLR (\( \dlrloss \)})}
    & \multicolumn{2}{|c|}{\textbf{GAMA\_PGD}}
    & \multicolumn{2}{|c|}{\textbf{MIFPE (\( \lnsceloss \))}}
    & \multicolumn{2}{|c|}{\textbf{T-MIFPE (\( \lnscelossyu \))}}
    & \thead{AA}\\

& \multicolumn{1}{|c|}{\textbf{}}
& \multicolumn{1}{|c|}{\textbf{}}
    & \thead{100}
    & \multicolumn{2}{|c}{\textbf{100}}
    & \multicolumn{2}{|c}{\textbf{100}}
    & \multicolumn{2}{|c|}{\textbf{100}}
    & \multicolumn{2}{|c|}{\textbf{100}}
    & \multicolumn{2}{|c|}{\textbf{100}}
    & \thead{4900}\\
\midrule
\multicolumn{13}{c}{%
    \textbf{\mnist}, \( \ell_{\infty} \),
    \( \varepsilon =0.3 \)} \\
\midrule
Uncovering limits~\cite{gowal2020uncovering} 
    & WRN-28-10
    & 99.26 & 96.55
    & 96.64 & (+0.09)
    & 96.71 & (+0.16) 
    & 96.69 & (+0.14)
    & 96.53 & \tbgreen{(-0.02)}
    & \tbnum{ 96.51} & \tbgreen{($\bm{-0.04 }$)}
    & 96.31 \\
MMA training~\cite{ding20mma}
    & LeNet5Madry
    & 98.95 & 95.09
    & 95.33 & (+0.24)
    & 95.59 & (+0.50)
    & 95.74 & (+0.19)
    & 94.91 & \tbgreen{(-0.18)} 
    & \tbnum{94.80} & \tbgreen{($\bm{-0.29 }$)} 
    & 91.40\\
TRADES~\cite{zhang19trades}
    & SmallCNN
    & 99.48 & 93.69
    & 93.88 & (+0.19)
    & 94.49 & (+0.80)
    & 93.82 & (+0.13)
    & 93.63 & \tbgreen{(-0.06)}
    & \tbnum{93.62} & \tbgreen{($\bm{-0.07}$)}
    & 92.71 \\
Robust optimization~\cite{madry18} 
    & SmallCNN
    & 99.35 & 93.06
    & 93.19 & (+0.13)
    & 93.63 & (+0.57)
    & 93.39 & (+0.33)
    & 93.08 & \tbgreen{(-0.18)} 
    & \tbnum{93.07} & \tbgreen{($\bm{-0.19 }$)} 
    & 90.85 \\
Fast adversarial training~\cite{wong20fast}
    & SmallCNN
    & 98.50 & 86.82
    & 86.96 & (+0.14)
    & 87.42 & (+0.60)
    & 87.62 & (+0.80)
    & 86.60 & \tbgreen{(-0.22)}
    & \tbnum{86.59} & \tbgreen{($\bm{-0.23}$)}
    & 82.93 \\
\midrule
\multicolumn{13}{c}{%
    \textbf{\cifarx}, \( \ell_{\infty} \),
    \( \varepsilon =8/255 \)} \\
\midrule
Uncovering limits~\cite{gowal2020uncovering} 
    & WRN-70-16
    & 91.10 & 67.96
    & 66.70 & \tbgreen{(-1.26)}
    & 66.78 & \tbgreen{(-1.18)}  
    & 66.08 & \tbgreen{(-1.88)}
    & 65.96 & \tbgreen{(-2.00)}
    & \tbnum{65.95} & \tbgreen{($\bm{-2.01}$)}
    & 65.87\\
Fixing data augmentation~\cite{rebuffi2021data} 
    & WRN-106-16
    & 88.50 & 67.57
    & 65.55 & \tbgreen{(-2.02)}
    & 65.61 & \tbgreen{(-1.96)}
    & 64.94 & \tbgreen{(-2.63)}
    & 64.78 & \tbgreen{(-2.79)}
    & \tbnum{ 64.72} & \tbgreen{($\bm{ -2.85}$)} 
    & 64.58 \\
Fixing data augmentation~\cite{rebuffi2021data} 
    & WRN-70-16
    & 88.54 & 67.27
    & 65.23 & \tbgreen{(-2.04)}
    & 65.32 & \tbgreen{(-1.95)}
    & 64.57 & \tbgreen{(-2.70)}
    & 64.57 & \tbgreen{(-2.80)}
    & \tbnum{64.46} & \tbgreen{($\bm{-2.81 }$)}
    & 64.20 \\
Uncovering limits~\cite{gowal2020uncovering} 
    & WRN-28-10
    & 89.48 & 65.59
    & 63.62 & \tbgreen{(-1.97)}
    & 63.82 & \tbgreen{(-1.77)}
    & 63.05 & \tbgreen{(-2.90)}
    & 62.97 & \tbgreen{(-2.62)}
    & \tbnum{62.94} & \tbgreen{($\bm{-2.65}$)}
    & 62.76 \\
Adversarial weight perturbation~\cite{wu2021wider}
    & WRN-28-10
    & 88.25 & 63.18
    & 60.51 & \tbgreen{(-2.67)}
    & 60.60 & \tbgreen{(-2.58)} 
    & 60.18 & \tbgreen{(-3.00)}
    & 60.10 & \tbgreen{(-3.08)}
    & \tbnum{60.09} & \tbgreen{($\bm{-3.09}$)}
    & 60.04\\
Unlabeled data~\cite{carmon2019unlabeled}
    & WRN-28-10
    & 89.69 & 61.60
    & 60.47 & \tbgreen{(-1.13)}
    & 60.67 & \tbgreen{(-0.93)}
    & 59.82 & \tbgreen{(-1.78)}
    & 59.73 & \tbgreen{(-1.87)} 
    & \tbnum{59.70} & \tbgreen{($\bm{-1.90}$)} 
    & 59.53 \\
HYDRA~\cite{sehwag2020hydra}
    & WRN-28-10
    & 88.98 & 59.53
    & 58.21 & \tbgreen{(-1.32)}
    & 58.30 & \tbgreen{(-1.23)}
    & 57.52 & \tbgreen{(-2.01)}
    & 57.39 & \tbgreen{(-2.14)}
    & \tbnum{57.36} & \tbgreen{($\bm{-2.17}$)} 
    & 57.14 \\
Pre-training~\cite{hendrycks2019using}
    & WRN-28-10
    & 87.11 & 57.07
    & 56.27 & \tbgreen{(-0.80)}
    & 57.07 & (0.00)
    & 55.22 & \tbgreen{(-1.85)}
    & 55.12 & \tbgreen{(-1.95)}
    & \tbnum{55.09} & \tbgreen{($\bm{-1.98}$)}
    & 54.92\\
Overfitting~\cite{rice20overfitting}
    & WRN-34-20
    & 85.34 & 56.85
    & 55.22 & \tbgreen{(-1.63)}
    & 55.97 & \tbgreen{(-0.88)}
    & 53.87 & \tbgreen{(-2.98)}
    & 53.67 & \tbgreen{(-3.18)}
    & \tbnum{53.66} & \tbgreen{($\bm{-3.19 }$)}
    & 53.42 \\
Self-adaptive training~\cite{huang20selfadaptive}\supddagger{}
    & WRN-34-10
    & 83.48 & 56.12
    & 54.30 & \tbgreen{(-1.82)}
    & 54.73 & \tbgreen{(-1.39)}
    & 53.64 & \tbgreen{(-2.48)}
    & 53.652 & \tbgreen{(-2.60)}
    & \tbnum{ 53.51 } & \tbgreen{($\bm{ -2.61 }$)}
    & 53.34\\
TRADES~\cite{zhang19trades}\supddagger{}
    & WRN-34-10
    & 84.92 & 55.21
    & 53.94 & \tbgreen{(-1.27)}
    & 54.11 & \tbgreen{(-1.10)}
    & 53.38 & \tbgreen{(-1.83)}
    & \tbnum{53.27} & \tbgreen{($\bm{-1.94}$)} 
    & \tbnum{53.27} & \tbgreen{($\bm{-1.94 }$)} 
    & 53.08\\
Neural level sets~\cite{atzmon19levelsets}\supddagger{}
    & RN-18
    & 81.30 & 79.12
    & 40.07 & \tbgreen{(-39.05)}
    & 45.10 & \tbgreen{(-34.02)}
    & 79.69 & (+0.57)
    & 40.15 & \tbgreen{(-38.97)}
    & \tbnum{40.12} & \tbgreen{($\bm{ -39.00}$)}
    & 39.77\\
MMA training~\cite{ding20mma}
    & WRN-28-10
    & 83.28 & 47.69
    & 48.66 & (+0.97)
    & 48.69 & (+1.00)
    & 48.07 & (+0.38)
    & 47.24 & \tbgreen{(-0.45)}
    & \tbnum{47.05} & \tbgreen{($\bm{-0.64}$)}
    & 39.76\\
YOPO~\cite{zhang19yopo}
    & WRN-34-10
    & 87.20 & 46.05
    & 47.02 & (+0.97)
    & 47.55 & (+1.50) 
    & 45.30 & \tbgreen{(-0.75)}
    & \tbnum{45.08} & \tbgreen{($\bm{-0.97}$)}
    & \tbnum{45.08} & \tbgreen{($\bm{-0.97}$)}
    & 44.83\\
Fast adversarial training~\cite{wong20fast}
    & RN-18
    & 83.34 & 45.75
    & 45.81 & (+0.06)
    & 46.89 & (+1.14) 
    & 43.71 & \tbgreen{(-2.04)}
    & \tbnum{43.57} & \tbgreen{($\bm{-2.18}$)}
    & \tbnum{43.57} & \tbgreen{($\bm{-2.18}$)}
    & 43.21 \\
\midrule
\multicolumn{13}{c}{%
    \textbf{\cifarc}, \( \ell_\infty \),
    \( \epsilon = 8/255 \)
} \\
\midrule
Adversarial weight perturbation~\cite{wu20wp}
    & WRN-34-10
    & 60.38 & 33.09
    & 30.74 & \tbgreen{(-2.35)}
    & 31.13 & \tbgreen{(-1.96)}
    & 29.44 & \tbgreen{(-3.65)}
    & 29.32 & \tbgreen{(-3.77)} 
    & \tbnum{29.28} & \tbgreen{($\bm{-3.81}$)} 
    & 28.86\\
Progressive Hardening~\cite{sitawarin20ates}
    & WRN-34-10
    & 62.82 & 26.18
    & 26.69 & (+0.51)
    & 27.26 & (+1.08)
    & 24.97 & \tbgreen{(-1.21)}
    & 25.24 & \tbgreen{( 0.94)} 
    & \tbnum{25.16} & \tbgreen{($\bm{-1.02 }$)} 
    & 24.57\\
Overfitting~\cite{rice20overfitting}
    & RN-18
    & 53.83 & 20.47
    & 20.17 & \tbgreen{(-0.30)}
    & 20.30 & \tbgreen{(-0.17)} 
    & 19.20 & \tbgreen{(-1.27)}
    & 19.40 & \tbgreen{(-1.07)}
    & \tbnum{19.34} & \tbgreen{($\bm{-1.13 }$)}
    & 18.95\\
\midrule
\multicolumn{13}{c}{%
    \textbf{\imagenet}, \( \ell_\infty \),
    \( \epsilon = 4/255 \)
} \\
\midrule
Robustness library~\cite{engstrom19}
    & RN-50
    & 62.56 & 32.16
    & 32.24 & (+0.08)
    & 32.80 & (+0.64)
    & 29.72 & \tbgreen{(-2.44)}
    & \tbnum{30.06} & \tbgreen{($\bm{-2.10}$)}
    & 30.08 & \tbgreen{(-2.08)}
    & 29.22\\
Transfer Better~\cite{salman2020adversarially}
    & RN-18
    & 52.92 & 29.30
    & 27.14 & \tbgreen{(-2.16)}
    & 27.40 & \tbgreen{(1.90)} 
    & 25.56 & \tbgreen{(-3.74)}
    & \tbnum{25.64} & \tbgreen{($\bm{ -3.66}$)}
    & 25.72 & \tbgreen{(-3.58)}
    & 18.95\\
\midrule
\multicolumn{13}{c}{%
    \textbf{\cifarx}, \( \ell_{2} \),
    \( \varepsilon = 0.5 \)} \\
\midrule
Uncovering limits~\cite{gowal2020uncovering}
    & WRN-70-16
    & 94.74 & 81.71
    & 80.93 &  \tbgreen{(-0.78)}
    & 80.94 &  \tbgreen{(-0.77)} 
    & 87.74 & (+6.03)
    & \tbnum{80.57} & \tbgreen{($\bm{-1.14}$)}
    & \tbnum{80.57} & \tbgreen{($\bm{-1.14}$)}
    & 80.53\\
Uncovering limits~\cite{gowal2020uncovering}
    & WRN-70-16
    & 90.90 & 75.20
    & 74.89 & \tbgreen{(-0.31)}
    & 74.95 & \tbgreen{(-0.25)}
    & 81.78 & (+6.58)
    & \tbnum{74.56} & \tbgreen{($\bm{-0.64}$)} 
    & \tbnum{74.56} & \tbgreen{($\bm{-0.64}$)} 
    & 74.50\\
Adversarial weight perturbation~\cite{wu20wp}
    & WRN-34-10
    & 88.51 & 74.77
    & 73.88 & \tbgreen{(-0.89)}
    & 73.89 & \tbgreen{(-0.88)}
    & 79.38 & (+4.61)
    & 73.67 & \tbgreen{(-1.10)}
    & \tbnum{73.66} & \tbgreen{($\bm{-1.11}$)}
    & 73.66 \\
Decoupling direction and norm~\cite{rony2019decoupling}
    & WRN-28-10
    & 89.05 & 66.56
    & 67.00 & (+0.44)
    & 67.02 & (+0.46)
    & 74.61 & (+8.05)
    & 66.50 & \tbgreen{(-0.06)}
    & \tbnum{66.49} & \tbgreen{($\bm{-0.07 }$)} 
    & 66.44\\
MMA training~\cite{ding20mma}
    & WRN-28-4
    & 88.02 & 66.22
    & 66.58 & (+0.36)
    & 66.60 & (+0.38)
    & 71.44 & (+5.22)
    & \tbnum{66.16} & \tbgreen{($\bm{-0.06}$)}
    & \tbnum{66.16} & \tbgreen{($\bm{-0.06}$)}
    & 66.09\\
\bottomrule
\end{tabular}}
\end{table*}

In ~\Cref{tab:loss}, we present a comprehensive analysis demonstrating that our proposed \(\yuloss\) loss function achieves substantial and consistent improvements over CE, C\&W, DLR, and the GAMA-PGD attack across all evaluated models, building on the robust framework of MIFPE~\cite{yu2023efficient} with refined enhancements. Unlike competing loss functions, which, as indicated by numbers in parentheses reflecting improvements relative to the CE baseline, often exhibit degraded performance on certain models (denoted by non-green numbers in parentheses), \(\yuloss\) and MIFPE consistently surpass CE. This uniform enhancement highlights a core deficiency in CE: relative errors in gradient computations induced by floating-point arithmetic. When combined with PGD, utilizing momentum and a step-size schedule, \(\yuloss\) achieves near-optimal robustness in only 100 iterations, closely approximating RobustBench~\cite{croce2020robustbench} benchmarks derived from over 4900 iterations. For example, on Uncovering limits~\cite{gowal2020uncovering} for CIFAR-10 (\( \ell_\infty \), \( \varepsilon = 8/255 \)), \(\yuloss\) reaches 65.95\%, compared to 65.87\% via AA, a mere 0.07 difference. Although \(\yuloss\)’s gains over MIFPE are modest, they are significant within the narrow scope for improvement, as both methods approach the upper bounds of attack efficacy in limited iterations. Uniquely, \(\yuloss\) leverages a theoretical foundation that minimizes gradient errors from floating-point operations, providing a principled advancement over prior methods.

\section{Conclusion}\label{sec:conclusion}

This work pioneers a comprehensive theoretical framework that systematically dissects floating-point-induced relative errors in gradient computations for adversarial attacks. By conducting the first in-depth analysis of numerical errors across four distinct attack scenarios---(i) unsuccessful untargeted attacks, (ii) successful untargeted attacks, (iii) unsuccessful targeted attacks, and (iv) successful targeted attacks---we uncover novel patterns of gradient computation instability, establishing robust theoretical foundations for understanding numerical error behavior. Leveraging these insights, we introduce the Theoretical Minimize the Impact of Floating Point Error (\(\yuloss\)) loss function, which incorporates an optimal scaling factor \( T = t^* \) to minimize floating-point error impact, significantly enhancing gradient accuracy. Extensive experiments on the MNIST, CIFAR-10 and CIFAR-100 datasets demonstrate that \(\yuloss\) outperforms existing loss functions, including CE, C\&W, DLR, and MIFPE, in attack potency and robustness evaluation accuracy. Our contributions not only advance the theoretical understanding of numerical stability in gradient-based attacks but also provide a generalizable methodology for designing numerically robust loss functions, paving the way for more reliable adversarial evaluation and model robustness assessment.

\clearpage

{\small
\bibliography{references}
}
\clearpage
\section{Technical Appendices and Supplementary Material}

\subsection{Successful Attack Phase of Untargeted Attacks}\label{sec:method_untargeted_success}
{  When \(\mathbf{z}_y \neq \mathbf{z}_{\pi_1}\),}Here, we assume \( \mathbf{z}_y = \mathbf{z}_{\pi_j} \) for some \( j \in \{2, \ldots, K\} \).

\begin{equation}
CE\left( \mathbf{z,}y \right) =-\log p_y=-\log \frac{e^{\mathbf{z}_y-\mathbf{z}_{\pi 2}}}{\sum_{i=1}^K{e^{\mathbf{z}_i-\mathbf{z}_{\pi 2}}}}
\label{eq:SCE_u_s}
\end{equation}

\begin{equation}
CE\left( c\mathbf{z},y \right) =-\log p_y^{c}=-\log \frac{e^{c(\mathbf{z}_y-\mathbf{z}_{\pi_2})}}{\sum_{i=1}^K{e^{c(\mathbf{z}_i-\mathbf{z}_{\pi_2})}}}
\label{eq:SCE_c_u_s}
\end{equation}

\begin{equation}
\begin{split}
 \nabla _{\mathbf{\hat{x}}}CE\left( c\mathbf{z,}y \right) & = c\left( -1+p_{y}^{c} \right) \nabla _{\mathbf{\hat{x}}}\left( \mathbf{z}_y-\mathbf{z}_{\pi 2} \right)  
 +\sum_{i\ne y}{cp_{i}^{c}\nabla _{\mathbf{\hat{x}}}\left( \mathbf{z}_i-\mathbf{z}_{\pi 2} \right)}\\
& = c\left( -1+p_{\pi_j}^{c} \right) \nabla _{\mathbf{\hat{x}}}\left( \mathbf{z}_{\pi_j}-\mathbf{z}_{\pi 2} \right) +\sum_{i\ne j} {cp_{\pi_i}^{c}\nabla _{\mathbf{\hat{x}}}\left( \mathbf{z}_{\pi_i}-\mathbf{z}_{\pi 2} \right)}\\
& = c\left( -1+p_{\pi_j}^{c} \right) \nabla _{\mathbf{\hat{x}}}\left( \mathbf{z}_{\pi_j}-\mathbf{z}_{\pi_{j-1}} + ... + \mathbf{z}_{\pi_3} - \mathbf{z}_{\pi 2} \right) \\ 
&  \quad + c  p_{\pi_{1}}^{c} \nabla _{\mathbf{\hat{x}}} \left(  \mathbf{z}_{\pi_{1}} -  \mathbf{z}_{\pi_{2}} \right)  \\
 &\quad + c  p_{\pi_{3}}^{c} \nabla _{\mathbf{\hat{x}}} \left( \mathbf{z}_{\pi_{3}} -\mathbf{z}_{\pi_{2}} \right) \\
 & \quad + ...\\
 &\quad + c  p_{\pi_{K}}^{c} \nabla _{\mathbf{\hat{x}}} \left( \mathbf{z}_{\pi_{K}} -\mathbf{z}_{\pi_{K-1}}+ ...+ \mathbf{z}_{\pi_{3}} -  \mathbf{z}_{\pi_{2}} \right) \\
&=c p_{\pi_{1}}^{c}\nabla _{\mathbf{\hat{x}}}\left( \mathbf{z}_{\pi_{1}}-\mathbf{z}_{\pi_{2}} \right)  \\
&\quad + c(1-p_{\pi_{j}}^{c}-p_{\pi_{3}}^{c}- ...  -p_{\pi_{K}}^{c} )\nabla _{\mathbf{\hat{x}}}\left( \mathbf{z}_{\pi_{2}}-\mathbf{z}_{\pi_{3}} \right)  \\
& \quad + ... \\
&\quad + c(1-p_{\pi_{j}}^{c}-p_{\pi_{j}}^{c}-...-p_{\pi_{K}}^{c})\nabla _{\mathbf{\hat{x}}}\left( \mathbf{z}_{\pi_{j-1}}-\mathbf{z}_{\pi_{j}} \right)  \\
&\quad + c(p_{\pi_{j+1}}^{c}+ ... + p_{\pi_{K}}^{c} )\nabla _{\mathbf{\hat{x}}}\left( \mathbf{z}_{\pi_{j+1}}-\mathbf{z}_{\pi_{j}} \right)  \\
& \quad + ... \\
&\quad + cp_{\pi_{K}}^{c}\nabla _{\mathbf{\hat{x}}}\left( \mathbf{z}_{\pi_{K}}-\mathbf{z}_{\pi_{K-1}} \right)  \\
&=c p_{\pi_{1}}^{c}\nabla _{\mathbf{\hat{x}}}\left( \mathbf{z}_{\pi_{1}}-\mathbf{z}_{\pi_{2}} \right)  \\
&\quad + c(p_{\pi_{1}}^{c}+p_{\pi_{2}}^{c} -p_{\pi_{j}}^{c})\nabla _{\mathbf{\hat{x}}}\left( \mathbf{z}_{\pi_{2}}-\mathbf{z}_{\pi_{3}} \right)  \\
& \quad + ... \\
&\quad + c(p_{\pi_{1}}^{c}+...+p_{\pi_{j-1}}^{c}-p_{\pi_{j}}^{c})\nabla _{\mathbf{\hat{x}}}\left( \mathbf{z}_{\pi_{j-1}}-\mathbf{z}_{\pi_{j}} \right)  \\
&\quad + c(p_{\pi_{j+1}}^{c}+ ... + p_{\pi_{K}}^{c} )\nabla _{\mathbf{\hat{x}}}\left( \mathbf{z}_{\pi_{j+1}}-\mathbf{z}_{\pi_{j}} \right)  \\
& \quad + ... \\
&\quad + cp_{\pi_{K}}^{c}\nabla _{\mathbf{\hat{x}}}\left( \mathbf{z}_{\pi_{K}}-\mathbf{z}_{\pi_{K-1}} \right)  \\
\end{split}
\end{equation}

where \(c=\frac{t}{\Delta_{\text{value}} }\) is a scale factor, \(t>0\), and $p_{\pi_i}^{c}=e^{c\left( \mathbf{z}_{\pi_i}-\mathbf{z}_{\pi 2} \right)}/\sum_{j=1}^K{e^{c\left( \mathbf{z}_{\pi_j}-\mathbf{z}_{\pi 2} \right)}}$.

\begin{equation}
\begin{split}
 \max_{i \neq y} \mathbf{z}_i -\mathbf{z}_y   =   \mathbf{z}_{\pi_1} -\mathbf{z}_{\pi_j}   =   (\mathbf{z}_{\pi_1} - \mathbf{z}_{\pi_2}) + ... + (\mathbf{z}_{\pi_{j-1}} - \mathbf{z}_{\pi_j}) 
\label{eq:untargeted_obj_success}
\end{split}
\end{equation}

Based on the primary objective of the untargeted attack in the Successful Attack Phase, as defined in Equation~\eqref{eq:untargeted_obj_success}, the gradients \( \nabla_{\mathbf{\hat{x}}} (\mathbf{z}_{\pi_1} - \mathbf{z}_{\pi_2}) \), \( \nabla_{\mathbf{\hat{x}}} (\mathbf{z}_{\pi_2} - \mathbf{z}_{\pi_3}) \), ..., \( \nabla_{\mathbf{\hat{x}}} (\mathbf{z}_{\pi_{j-1}} - \mathbf{z}_{\pi_j}) \) are critical components. To enhance the accuracy of gradient computations in such attacks, it is necessary to simultaneously minimize the upper bounds of the relative errors associated with the terms \( |c p_{\pi_1}^c \nabla_{\mathbf{\hat{x}}} (\mathbf{z}_{\pi_1} - \mathbf{z}_{\pi_2})| \), \( |c (p_{\pi_1}^c + p_{\pi_2}^c - p_{\pi_j}^c) \nabla_{\mathbf{\hat{x}}} (\mathbf{z}_{\pi_2} - \mathbf{z}_{\pi_3})| \), ..., \( |c (p_{\pi_1}^c + \dots + p_{\pi_{j-1}}^c - p_{\pi_j}^c) \nabla_{\mathbf{\hat{x}}} (\mathbf{z}_{\pi_{j-1}} - \mathbf{z}_{\pi_j})| \). We define \( \delta(t)_{u\_s\_i-1\_i}^{\text{sup}\_\text{min}} \) as the upper bound of the relative error in \( \nabla_{\mathbf{\hat{x}}} (\mathbf{z}_{\pi_{i-1}} - \mathbf{z}_{\pi_i}) \) during the Successful Attack Phase of an untargeted attack, where \( i \) ranges from 2 to \( j \), and \(j \in \{2,...,K\}\):

When $  i=2$:

\begin{equation}
\delta(t)_{u\_s\_1\_2}^{\text{sup}\_\text{min}} = \frac{\epsilon_{\text{max}}}{|c p_{\pi_1}^c \nabla_{\mathbf{\hat{x}}} (\mathbf{z}_{\pi_1} - \mathbf{z}_{\pi_2})|_{\text{max}}}
\end{equation}

When $j\geq i>2$:

\begin{equation}
\begin{split}
&\delta(t)_{u\_s\_i-1\_i}^{\text{sup}\_\text{min}} \\
&= \frac{\epsilon_{\text{max}}}{|c (p_{\pi_1}^c + \dots + p_{\pi_{i-1}}^c - p_{\pi_j}^c) \nabla_{\mathbf{\hat{x}}} (\mathbf{z}_{\pi_{i-1}} - \mathbf{z}_{\pi_i})|_{\text{max}}} \\
&< \frac{\epsilon_{\text{max}}}{|c p_{\pi_1}^c \nabla_{\mathbf{\hat{x}}} (\mathbf{z}_{\pi_{i-1}} - \mathbf{z}_{\pi_i})|_{\text{max}}}
\end{split}
\end{equation}

It is evident that \( c p_{\pi_1}^c \) is a pivotal factor in controlling the upper bounds of all terms from \( \delta(t)_{u\_s\_1\_2}^{\text{sup}\_\text{min}} \) to \( \delta(t)_{u\_s\_j-1\_j}^{\text{sup}\_\text{min}} \). Increasing the value of \( c p_{\pi_1}^c \) effectively reduces these upper bounds. We define \( g(t)_{u\_s} = c p_{\pi_1}^c \), and its derivative is given by:

\begin{equation}
g'(t)_{u\_s} = \frac{p_{\pi_1}^{c}(B+c (\Delta_{\text{value}}B -S) )}{\Delta_{\text{value}} B} > 0,
\end{equation}

where,
$\Delta_{\text{value}} = \mathbf{z}_{\pi_1} - \mathbf{z}_{\pi_2} ,
 B = \sum_{j=1}^K e^{c(\mathbf{z}_j - \mathbf{z}_{\pi_2})} = 1 + \sum_{j\neq\pi_2} e^{c(\mathbf{z}_j - \mathbf{z}_{\pi_2})} ,
  S = \sum_{j=1}^K (\mathbf{z}_j - \mathbf{z}_{\pi_2})e^{c(\mathbf{z}_j - \mathbf{z}_{\pi_2})} ,
  \Delta_{\text{value}}B - S = \sum_{j=1}^K (\mathbf{z}_{\pi_1} - \mathbf{z}_{\pi_j})e^{c(\mathbf{z}_j - \mathbf{z}_{\pi_2})} >0$.


This indicates that \( g(t)_{u\_s} \) is monotonically increasing. As \( t \) increases, \( g(t)_{u\_s} \) grows accordingly; however, \( t \) is constrained by floating-point underflow. We define \( \lambda > 0 \) as the threshold beyond which \( e^{-\lambda} = 0 \) due to underflow, with \( \lambda \) taking values of 16.6355, 103.2789, and 744.4401 for 16-bit, 32-bit, and 64-bit floating-point representations, respectively. When \( t (\mathbf{z}_{\pi_j} - \mathbf{z}_{\pi_1}) / (\mathbf{z}_{\pi_1} - \mathbf{z}_{\pi_2}) < -\lambda \), \( p_{\pi_j}^c \) becomes zero due to underflow. Notably, smaller values of \( t \) introduce larger relative errors in gradient computations. To mitigate this, we constrain the minimum value of \( t^{*} \) to be no less than 1. To ensure \( p_{\pi_j}^c \) remains non-zero during the attack process, the maximum value of \( t^{*} \) is bounded by \( \frac{\lambda (\mathbf{z}_{\pi_1} - \mathbf{z}_{\pi_2})}{\mathbf{z}_{\pi_1} - \mathbf{z}_{\pi_j}} \). Thus, \( t^{*} \) is defined as:
\begin{equation}
t^{*} = \max\left\{1, \frac{\lambda (\mathbf{z}_{\pi_1} - \mathbf{z}_{\pi_2})}{\mathbf{z}_{\pi_1} - \mathbf{z}_{\pi_j}}\right\}.
\label{eq:increase_t_star}
\end{equation}





\subsection{Unsuccessful Attack Phase of Targeted Attacks}\label{sec:method_targeted_unsuccess}
{When \(\mathbf{z}_{y_t} \neq \mathbf{z}_{\pi_1}\),}
Here, we assume \( \mathbf{z}_{y_t} = \mathbf{z}_{\pi_j} \) for some \( j \in \{2, \ldots, K\} \).

\begin{equation}
-CE\left( \mathbf{z}, y_t \right) =\log p_{y_t}=\log \frac{e^{\mathbf{z}_{y_t}-\mathbf{z}_{\pi_1}}}{\sum_{i=1}^K{e^{\mathbf{z}_i-\mathbf{z}_{\pi_1}}}}
\label{eq:SCE_targeted_u}
\end{equation}

\begin{equation}
-CE\left( c\mathbf{z},y_t \right) =\log p_{y_t}^{c}=\log \frac{e^{c(\mathbf{z}_{y_t}-\mathbf{z}_{\pi_1})}}{\sum_{i=1}^K{e^{c(\mathbf{z}_i-\mathbf{z}_{\pi_1})}}}
\label{eq:SCE_c_targeted_u}
\end{equation}

\begin{equation}
\begin{split}
 -\nabla _{\mathbf{\hat{x}}}CE\left( c\mathbf{z,}y_t \right) 
&= c\left( 1 -p_{y_t}^{c} \right) \nabla _{\mathbf{\hat{x}}}\left( \mathbf{z}_{y_t} -\mathbf{z}_{\pi_1} \right)  -\sum_{i\ne y_t}{cp_{i}^{c}\nabla _{\mathbf{\hat{x}}}\left( \mathbf{z}_i-\mathbf{z}_{\pi_1} \right)}\\
&= c\left( 1 -p_{\pi_j}^{c} \right) \nabla _{\mathbf{\hat{x}}}\left( \mathbf{z}_{\pi_j} -\mathbf{z}_{\pi_1} \right)  -\sum_{i\ne j}{cp_{i}^{c}\nabla _{\mathbf{\hat{x}}}\left( \mathbf{z}_{\pi_i}-\mathbf{z}_{\pi_1} \right)}\\
& = c\left( 1 -p_{\pi_j}^{c} \right)\nabla _{\mathbf{\hat{x}}} \left( \mathbf{z}_{\pi_{j}} -\mathbf{z}_{\pi_{j-1}}+ ...+ \mathbf{z}_{\pi_{2}} -  \mathbf{z}_{\pi_{1}} \right) \\
&\quad - c  p_{\pi_{2}}^{c} \nabla _{\mathbf{\hat{x}}} \left(  \mathbf{z}_{\pi_{2}} -  \mathbf{z}_{\pi_{1}} \right)  \\
&\quad - c  p_{\pi_{3}}^{c} \nabla _{\mathbf{\hat{x}}} \left( \mathbf{z}_{\pi_{3}} -\mathbf{z}_{\pi_{2}}+  \mathbf{z}_{\pi_{2}} -  \mathbf{z}_{\pi_{1}} \right) \\
& \quad - ...\\
&\quad - c  p_{\pi_{K}}^{c} \nabla _{\mathbf{\hat{x}}} \left( \mathbf{z}_{\pi_{K}} -\mathbf{z}_{\pi_{K-1}}+ ...+ \mathbf{z}_{\pi_{2}} -  \mathbf{z}_{\pi_{1}} \right) \\
&=c(1-p_{\pi_{j}}^{c}-p_{\pi_{2}}^{c}- ...  -p_{\pi_{K}}^{c} ) \nabla _{\mathbf{\hat{x}}}\left( \mathbf{z}_{\pi_{2}}-\mathbf{z}_{\pi_{1}} \right)  \\
&\quad + c(1-p_{\pi_{j}}^{c}-p_{\pi_{3}}^{c}- ...  -p_{\pi_{K}}^{c} )\nabla _{\mathbf{\hat{x}}}\left( \mathbf{z}_{\pi_{3}}-\mathbf{z}_{\pi_{2}} \right)  \\
& \quad + ... \\
&\quad + c(1-p_{\pi_{j}}^{c}-p_{\pi_{j}}^{c}-...-p_{\pi_{K}}^{c})\nabla _{\mathbf{\hat{x}}}\left( \mathbf{z}_{\pi_{j}}-\mathbf{z}_{\pi_{j-1}} \right)  \\
&\quad - c(p_{\pi_{j+1}}^{c}+ ... + p_{\pi_{K}}^{c} )\nabla _{\mathbf{\hat{x}}}\left( \mathbf{z}_{\pi_{j+1}}-\mathbf{z}_{\pi_{j}} \right)  \\
& \quad - ... \\
&\quad - cp_{\pi_{K}}^{c}\nabla _{\mathbf{\hat{x}}}\left( \mathbf{z}_{\pi_{K}}-\mathbf{z}_{\pi_{K-1}} \right)  \\
&=c(p_{\pi_{1}}^{c}-p_{\pi_{j}}^{c} ) \nabla _{\mathbf{\hat{x}}}\left( \mathbf{z}_{\pi_{2}}-\mathbf{z}_{\pi_{1}} \right)  \\
&\quad + c(p_{\pi_{1}}^{c} + p_{\pi_{2}}^{c}-p_{\pi_{j}}^{c} )\nabla _{\mathbf{\hat{x}}}\left( \mathbf{z}_{\pi_{3}}-\mathbf{z}_{\pi_{2}} \right)  \\
& \quad + ... \\
&\quad + c(p_{\pi_{1}}^{c}+...+p_{\pi_{j-1}}^{c}- p_{\pi_{j}}^{c})\nabla _{\mathbf{\hat{x}}}\left( \mathbf{z}_{\pi_{j}}-\mathbf{z}_{\pi_{j-1}} \right)  \\
&\quad - c(p_{\pi_{j+1}}^{c}+ ... + p_{\pi_{K}}^{c} )\nabla _{\mathbf{\hat{x}}}\left( \mathbf{z}_{\pi_{j+1}}-\mathbf{z}_{\pi_{j}} \right)  \\
& \quad - ... \\
&\quad - cp_{\pi_{K}}^{c}\nabla _{\mathbf{\hat{x}}}\left( \mathbf{z}_{\pi_{K}}-\mathbf{z}_{\pi_{K-1}} \right)  \\
\end{split}
\end{equation}
where \(c=\frac{t}{\Delta_{\text{value}} }\) is a scale factor, \(t>0\), \( \Delta_{\text{value}} = \mathbf{z}_{\pi_1} - \mathbf{z}_{\pi_2}>0\),  and $p_{\pi_j}^{c}=e^{c\left( \mathbf{z}_{\pi_j}-\mathbf{z}_{\pi_1} \right)}/\sum_{j=1}^K{e^{c\left( \mathbf{z}_{\pi_j}-\mathbf{z}_{\pi_1} \right)}}$.



\begin{equation}
\begin{split}
 \mathbf{z}_{y_t} - \max_{i \neq y_t} \mathbf{z}_i     =   \mathbf{z}_{\pi_j} -\mathbf{z}_{\pi_1}    =   (\mathbf{z}_{\pi_j} - \mathbf{z}_{\pi_{j-1}}) + ... + (\mathbf{z}_{\pi_{2}} - \mathbf{z}_{\pi_1}) 
\label{eq:targeted_obj_unsuccess}
\end{split}
\end{equation}

Based on the primary objective of the targeted attack in the Unsuccessful Attack Phase, as specified in Equation~\eqref{eq:targeted_obj_unsuccess}, the gradients \( \nabla_{\mathbf{\hat{x}}} (\mathbf{z}_{\pi_2} - \mathbf{z}_{\pi_1}) \), \( \nabla_{\mathbf{\hat{x}}} (\mathbf{z}_{\pi_3} - \mathbf{z}_{\pi_2}) \), ..., \( \nabla_{\mathbf{\hat{x}}} (\mathbf{z}_{\pi_{j}} - \mathbf{z}_{\pi_{j-1}}) \) are critical components for gradient computation accuracy.
To enhance the precision of these gradient computations, it is necessary to simultaneously minimize the upper bounds of the relative errors associated with the terms \( |c (p_{\pi_1}^c - p_{\pi_j}^c) \nabla_{\mathbf{\hat{x}}} (\mathbf{z}_{\pi_2} - \mathbf{z}_{\pi_1})| \), \( |c (p_{\pi_1}^c + p_{\pi_2}^c - p_{\pi_j}^c) \nabla_{\mathbf{\hat{x}}} (\mathbf{z}_{\pi_3} - \mathbf{z}_{\pi_2})| \), ..., \( |c (p_{\pi_1}^c + p_{\pi_2}^c + \dots + p_{\pi_{j-1}}^c - p_{\pi_j}^c) \nabla_{\mathbf{\hat{x}}} (\mathbf{z}_{\pi_j} - \mathbf{z}_{\pi_{j-1}})| \). We define \( \delta(t)_{t\_u\_i\_i-1}^{\text{sup}\_\text{min}} \) as the upper bound of the relative error in \( \nabla_{\mathbf{\hat{x}}} (\mathbf{z}_{\pi_i} - \mathbf{z}_{\pi_{i-1}}) \) during the Unsuccessful Attack Phase of a targeted attack, where \( i \) ranges from 2 to \( j \):

When $  i=2$:

\begin{equation}
\delta(t)_{t\_u\_2\_1}^{\text{sup}\_\text{min}} = \frac{\epsilon_{\text{max}}}{|c (p_{\pi_1}^c - p_{\pi_j}^c) \nabla_{\mathbf{\hat{x}}} (\mathbf{z}_{\pi_2} - \mathbf{z}_{\pi_1})|_{\text{max}}}
\end{equation}

When $j\geq i>2$:

\begin{equation}
\begin{split}
& \delta(t)_{t\_u\_i\_i-1}^{\text{sup}\_\text{min}} \\
&= \frac{\epsilon_{\text{max}}}{|c (p_{\pi_1}^c + p_{\pi_2}^c + \dots + p_{\pi_{i-1}}^c - p_{\pi_j}^c) \nabla_{\mathbf{\hat{x}}} (\mathbf{z}_{\pi_i} - \mathbf{z}_{\pi_{i-1}})|_{\text{max}}} \\
&< \frac{\epsilon_{\text{max}}}{|c (p_{\pi_1}^c - p_{\pi_j}^c) \nabla_{\mathbf{\hat{x}}} (\mathbf{z}_{\pi_i} - \mathbf{z}_{\pi_{i-1}})|_{\text{max}}}
\end{split}
\end{equation}

It is evident that \( c (p_{\pi_1}^c - p_{\pi_j}^c) \) is a pivotal factor in controlling the upper bounds of all terms from \( \delta(t)_{t\_u\_2\_1}^{\text{sup}\_\text{min}} \) to \( \delta(t)_{t\_u\_j\_j-1}^{\text{sup}\_\text{min}} \). Increasing the value of \( c (p_{\pi_1}^c - p_{\pi_j}^c) \) effectively reduces these upper bounds. We define \( g(t)_{t\_u} = c (p_{\pi_1}^c - p_{\pi_j}^c) \), with its derivative given by:

\begin{equation}
g'(t)_{t\_u} = \frac{A (B + c D) + c S}{\Delta_{\text{value}} A^2} > 0,
\end{equation}

where \( \Delta_{\text{value}} = \mathbf{z}_{\pi_1} - \mathbf{z}_{\pi_2} \), \( A = \sum_{i=1}^K e^{c (\mathbf{z}_{\pi_i} - \mathbf{z}_{\pi_1})} \), \( B = 1 - e^{c (\mathbf{z}_{\pi_j} - \mathbf{z}_{\pi_1})} > 0 \), \( D = - c e^{c (\mathbf{z}_{\pi_j} - \mathbf{z}_{\pi_1})} (\mathbf{z}_{\pi_j} - \mathbf{z}_{\pi_1}) > 0 \), and \( S = -\sum_{i=1}^K e^{c (\mathbf{z}_{\pi_i} - \mathbf{z}_{\pi_1})} (\mathbf{z}_{\pi_i} - \mathbf{z}_{\pi_1}) > 0 \). This indicates that \( g(t)_{t\_u} \) is monotonically increasing. As \( t \) increases, \( g(t)_{t\_u} \) grows accordingly; 
Similar to the analysis of $t^{*}$ in Section~\ref{sec:method_targeted_unsuccess}, we define $t^{*}$ according to the following relation:
\begin{equation}
t^{*} = \max\left\{1, \frac{\lambda (\mathbf{z}_{\pi_1} - \mathbf{z}_{\pi_2})}{\mathbf{z}_{\pi_1} - \mathbf{z}_{\pi_j}}\right\}
\end{equation}

\subsection{Successful Attack Phase of Targeted Attacks}\label{sec:method_targeted_success}
{ When \(\mathbf{z}_{y_t} = \mathbf{z}_{\pi_1}\)}

\begin{equation}
-CE\left( \mathbf{z}, y_t \right) =\log p_{y_t}=\log \frac{e^{\mathbf{z}_{y_t}-\mathbf{z}_{\pi_2}}}{\sum_{i=1}^K{e^{\mathbf{z}_i-\mathbf{z}_{\pi_2}}}}
\label{eq:SCE_targeted_s}
\end{equation}

\begin{equation}
-CE\left( c\mathbf{z},y_t \right) =\log p_{y_t}^{c}=\log \frac{e^{c(\mathbf{z}_{y_t}-\mathbf{z}_{\pi_2})}}{\sum_{i=1}^K{e^{c(\mathbf{z}_i-\mathbf{z}_{\pi_2})}}}
\label{eq:SCE_c_targeted_s}
\end{equation}

\begin{equation}
\begin{split}
 -\nabla _{\mathbf{\hat{x}}}CE\left( c\mathbf{z,}y_t \right) 
&= c\left( 1 -p_{\pi_1}^{c} \right) \nabla _{\mathbf{\hat{x}}}\left( \mathbf{z}_{\pi_1}-\mathbf{z}_{\pi_2} \right)   -\sum_{i\ne 1}{cp_{\pi_i}^{c}\nabla _{\mathbf{\hat{x}}}\left( \mathbf{z}_{\pi_i}-\mathbf{z}_{\pi_2} \right)}\\
&=(1-cp_{\pi_{1}}^{c})\nabla _{\mathbf{\hat{x}}}\left( \mathbf{z}_{\pi_{1}}-\mathbf{z}_{\pi_{2}} \right)  \\
&\quad - cp_{\pi_{3}}^{c}\nabla _{\mathbf{\hat{x}}}\left( \mathbf{z}_{\pi_{3}}-\mathbf{z}_{\pi_{2}} \right)  \\
&\quad - cp_{\pi_{4}}^{c} \nabla _{\mathbf{\hat{x}}}\left( \mathbf{z}_{\pi_{4}}-\mathbf{z}_{\pi_{3}} + \mathbf{z}_{\pi_{3}} - \mathbf{z}_{\pi_{2}} \right)  \\
& \quad - ... \\
&\quad - c p_{\pi_{K}}^{c}\nabla _{\mathbf{\hat{x}}}\left( \mathbf{z}_{\pi_{K}}-\mathbf{z}_{\pi_{K-1}} + ... + \mathbf{z}_{\pi_{3}} - \mathbf{z}_{\pi_{2}} \right)  \\
&=(1-cp_{\pi_{1}}^{c})\nabla _{\mathbf{\hat{x}}}\left( \mathbf{z}_{\pi_{1}}-\mathbf{z}_{\pi_{2}} \right)  \\
&\quad - c(p_{\pi_{3}}^{c} + ... + p_{\pi_{K}}^{c})\nabla _{\mathbf{\hat{x}}}\left( \mathbf{z}_{\pi_{3}}-\mathbf{z}_{\pi_{2}} \right)  \\
&\quad - c(p_{\pi_{4}}^{c} + ... + p_{\pi_{K}}^{c})\nabla _{\mathbf{\hat{x}}}\left( \mathbf{z}_{\pi_{4}}-\mathbf{z}_{\pi_{3}} \right)  \\
& \quad - ... \\
&\quad - c p_{\pi_{K}}^{c}\nabla _{\mathbf{\hat{x}}}\left( \mathbf{z}_{\pi_{K}}-\mathbf{z}_{\pi_{K-1}} \right)  \\
\end{split}
\end{equation}
where \(y_{\text{t}}\) is a predefined target class, \(y_{\text{t}} \in \{1, 2, \ldots, K\}\), and \(y_{\text{t}} \neq y\),  \(c=\frac{t}{\Delta_{\text{value}} }\) is a scale factor, \(t>0\), \( \Delta_{\text{value}} = |\mathbf{z}_{y_t} - \max_{i \neq y_t} \mathbf{z}_i |\),  and $p_{i}^{c}=e^{c\left( \mathbf{z}_i-\mathbf{z}_{y_t} \right)}/\sum_{j=1}^K{e^{c\left( \mathbf{z}_j-\mathbf{z}_{y_t} \right)}}$.

\begin{equation}
\begin{split}
 \mathbf{z}_{y_t} - \max_{i \neq y_t} \mathbf{z}_i    =   \mathbf{z}_{\pi_1} -\mathbf{z}_{\pi_2}  
\label{eq:targeted_obj_success}
\end{split}
\end{equation}

Following the targeted attack objective defined in Equation~\eqref{eq:targeted_obj_success}, we establish that the gradient component $\nabla_{\mathbf{\hat{x}}}(\mathbf{z}_{\pi_1} - \mathbf{z}_{\pi_2})$ plays a pivotal role in unsuccessful attack scenarios. To optimize gradient computation accuracy, we specifically minimize the relative error $\delta{( t)}_{t\_s}$ in the gradient magnitude $|c(1-p_{\pi_1}^c)\nabla_{\mathbf{\hat{x}}}(\mathbf{z}_{\pi_1}-\mathbf{z}_{\pi_2})|$. 

\begin{equation}
\begin{split}
 \delta{( t)}_{t\_s}^{\text{sup}\_\text{min}} 
 = \frac{\epsilon_{\text{max}}}{  {|c (1 - p_{\pi_1}^c) \nabla _{\mathbf{\hat{x}}}\left(  \mathbf{z}_{\pi_1} -\mathbf{z}_{\pi_2} \right)|}_{max} },
\end{split}
\end{equation}

The subsequent analysis follows the same methodology as \cref{sec:method_untargeted_unsuccess}
this optimization is equivalent to finding the maximum values for the corresponding coefficients $c(1-p_{\pi_1}^c)$, 

\begin{equation}
{g( t)}_{t\_s} = c \left( 1 - p_{\pi_1}^c \right)>0
\end{equation}

\begin{equation}
{g'(t)}_{t\_s} = \frac{B^2 - B + c S}{\Delta_{\text{value}} B^2}.
\end{equation}
where $p_{\pi_1}^c = \frac{e^{c \cdot 0}}{\sum_{j=1}^K e^{c (\mathbf{z}_j - \mathbf{z}_{\pi_1})}} = \frac{1}{B}, \quad B = \sum_{j=1}^K e^{c (\mathbf{z}_j - \mathbf{z}_{\pi_1})} = 1 + \sum_{j \neq {\pi_1}} e^{c (\mathbf{z}_j - \mathbf{z}_{\pi_1})}> 1$
, and $ S = \sum_{j=1}^K (\mathbf{z}_j - \mathbf{z}_{\pi_1}) e^{c (\mathbf{z}_j - \mathbf{z}_{\pi_1})}<0 $ , due to \(\mathbf{z}_j - \mathbf{z}_{\pi_1}<0\), \( c = t / \Delta_{\text{value}} \), and \( \Delta_{\text{value}} = \mathbf{z}_{\pi_1} - \mathbf{z}_{\pi_2}>0\).

We define \( t^* \) as the point where \( g'(t^*)_{t\_s} = 0 \), given by:
\begin{equation}
t^* = \frac{\Delta_{\text{value}} B (B - 1)}{-S},
\label{eq:t_star_t_s}
\end{equation}
Where \( \Delta_{\text{value}} > 0 \), \( B > 1 \), and \( S < 0 \), ensuring \( t^* > 0 \). At \( t = t^* \), \( g(t)_{t\_s} \) achieves its maximum value.

\section{Limitations}\label{sec:limit}
In the theoretical analysis, we comprehensively examined both untargeted and targeted adversarial attacks, covering both their unsuccessful and successful phases. However, in the experimental section, we only compared the performance of our proposed Theoretical MIFPE (\(\yuloss\)) loss function against other loss functions in the context of untargeted attacks, using the MNIST, CIFAR-10, and CIFAR-100 datasets. We did not conduct experimental comparisons for targeted attack scenarios.  Future work could explore these dimensions to ensure broader applicability.

\section{Compute Resources}
\label{sec:compute_resources}

To ensure reproducibility of the experimental results presented in Section~\ref{sec:results}, we provide a detailed description of the compute resources used for all experiments. All experiments were conducted on a single NVIDIA GeForce RTX 2080 Ti GPU with 11 GB of GDDR6 memory. The system was equipped with 32 GB of RAM and 1 TB of SSD storage, running on a Linux-based operating system (Ubuntu 20.04). The software environment included Python 3.8, PyTorch 1.9, and standard libraries for implementing the PGD attack framework and loss functions (CE, C\&W, DLR, MIFPE, and \(\yuloss\)).

Each experimental run, consisting of a PGD attack with 100 iterations on the MNIST, CIFAR-10, or CIFAR-100 datasets, was executed under \(\ell_\infty\)- or \(\ell_2\)-bounded threat models. The approximate execution time for a single run varied by dataset due to differences in image size and model complexity:
\begin{itemize}
    \item \textbf{MNIST}: Approximately 5--10 minutes per run for a single model and loss function combination.
    \item \textbf{CIFAR-10}: Approximately 15--60 minutes per run.
    \item \textbf{CIFAR-100}: Approximately 15--60 minutes per run.
\end{itemize}
These times account for the standardized configuration (100 iterations, momentum factor of 0.75, and a linearly decaying step-size schedule) and include data loading, model evaluation, and adversarial example generation. For each dataset and threat model, we conducted multiple runs to compare the five loss functions across different models, as detailed in Table~\ref{tab:loss}. The total compute time for the experiments reported in the paper is estimated at approximately 150--200 hours of GPU time, depending on the specific configurations and models tested.

\end{document}